\PassOptionsToPackage{unicode}{hyperref}
\PassOptionsToPackage{hyphens}{url}
\documentclass[
  11pt,
]{article}
\usepackage{xcolor}
\usepackage[margin=0.8in]{geometry}
\usepackage{amsmath,amssymb}
\usepackage{iftex}
\ifPDFTeX
  \usepackage[T1]{fontenc}
  \usepackage[utf8]{inputenc}
  \usepackage{textcomp} 
\else 
  \usepackage{unicode-math} 
  \defaultfontfeatures{Scale=MatchLowercase}
  \defaultfontfeatures[\rmfamily]{Ligatures=TeX,Scale=1}
\fi
\usepackage{lmodern}
\ifPDFTeX\else
\fi
\IfFileExists{upquote.sty}{\usepackage{upquote}}{}
\IfFileExists{microtype.sty}{
  \usepackage[]{microtype}
  \UseMicrotypeSet[protrusion]{basicmath} 
}{}
\usepackage{setspace}
\makeatletter
\@ifundefined{KOMAClassName}{
  \IfFileExists{parskip.sty}{%
    \usepackage{parskip}
  }{
    \setlength{\parindent}{0pt}
    \setlength{\parskip}{6pt plus 2pt minus 1pt}}
}{
  \KOMAoptions{parskip=half}}
\makeatother
\usepackage{color}
\usepackage{fancyvrb}

\DefineVerbatimEnvironment{Highlighting}{Verbatim}{commandchars=\\\{\}}
\newenvironment{Shaded}{}{}

\newcommand{\DataTypeTok}[1]{\textcolor[rgb]{0.56,0.13,0.00}{#1}}
\newcommand{\DecValTok}[1]{\textcolor[rgb]{0.25,0.63,0.44}{#1}}

\newcommand{\FunctionTok}[1]{\textcolor[rgb]{0.02,0.16,0.49}{#1}}

\newcommand{\OtherTok}[1]{\textcolor[rgb]{0.00,0.44,0.13}{#1}}

\newcommand{\StringTok}[1]{\textcolor[rgb]{0.25,0.44,0.63}{#1}}

\usepackage{longtable,booktabs,array}
\usepackage{calc} 
\usepackage{etoolbox}
\makeatletter
\patchcmd\longtable{\par}{\if@noskipsec\mbox{}\fi\par}{}{}
\makeatother
\IfFileExists{footnotehyper.sty}{\usepackage{footnotehyper}}{\usepackage{footnote}}
\makesavenoteenv{longtable}
\usepackage{graphicx}
\makeatletter
\newsavebox\pandoc@box
\newcommand*\pandocbounded[1]{
  \sbox\pandoc@box{#1}%
  \Gscale@div\@tempa{\textheight}{\dimexpr\ht\pandoc@box+\dp\pandoc@box\relax}%
  \Gscale@div\@tempb{\linewidth}{\wd\pandoc@box}%
  \ifdim\@tempb\p@<\@tempa\p@\let\@tempa\@tempb\fi
  \ifdim\@tempa\p@<\p@\scalebox{\@tempa}{\usebox\pandoc@box}%
  \else\usebox{\pandoc@box}%
  \fi%
}
\def\fps@figure{htbp}
\makeatother
\providecommand{\tightlist}{%
  \setlength{\itemsep}{0pt}\setlength{\parskip}{0pt}}
\usepackage{bookmark}
\IfFileExists{xurl.sty}{\usepackage{xurl}}{} 
\hypersetup{
  hidelinks,
  pdfcreator={LaTeX via pandoc}}

\author{}
\date{}

\begin{document}

\setstretch{1.1}
\section{Evolve: A Persistent Knowledge Lifecycle for Small Language
Models}\label{evolve-a-persistent-knowledge-lifecycle-for-small-language-models}

\textbf{Dikran Hovagimian} Independent Researcher dikran@hovagimian.com

Code and evaluation data: https://gitlab.com/dikran.hovagimian/evolve

April 2026

\begin{center}\rule{0.5\linewidth}{0.5pt}\end{center}

\subsection{Abstract}\label{abstract}

Evolve pairs a small local language model with a persistent,
teacher-compiled knowledge store --- refined through sleep consolidation
and usage-driven refresh --- to deliver substantial accuracy gains over
the model's parametric baseline while amortizing teacher costs through
cross-query knowledge reuse. Rather than retrieving document fragments
at query time, Evolve constructs a store of semantically coherent
sections compiled by teacher models at natural conceptual boundaries;
new sections are staged on acquisition, consolidated offline through
teacher-mediated merging, and refreshed inline when expired. A
2B-parameter local model handles classification and generation; large
teacher models are invoked only for knowledge operations.

Across 750 benchmark queries spanning custom specialist questions,
NaturalQuestions, and TriviaQA, the 2B model augmented by Evolve
improves from \textbf{20--33\% baseline accuracy to 60--84\%}
(+40--52pp) while reducing teacher invocations by over 50\% through
reuse. Post-consolidation compresses the knowledge store by 31--33.5\%
across three independent benchmarks while preserving accuracy;
section-based retrieval outperforms chunk-based retrieval by 5--9pp
across every lifecycle condition. The architecture supports two
generation modes over the same lifecycle --- \textbf{suppress} (strict
section-only grounding, auditable) and \textbf{augment}
(section-supplemented responses).

\begin{center}\rule{0.5\linewidth}{0.5pt}\end{center}

\subsection{1. Introduction}\label{introduction}

\subsubsection{1.1 The Problem with Parametric
Knowledge}\label{the-problem-with-parametric-knowledge}

Large Language Models encode vast factual knowledge within their
parameters during training, but this parametric knowledge has
fundamental limitations: it is frozen at the training cutoff, cannot be
selectively updated, decays unpredictably across domains, and produces
hallucination when the model generates plausible-sounding content that
has no factual basis. Retrieval-Augmented Generation (RAG) mitigates
hallucination by grounding generation in retrieved documents, but
conventional RAG systems face persistent challenges:

\begin{enumerate}
\def\labelenumi{\arabic{enumi}.}
\item
  \textbf{Chunk-boundary mismatch.} Fixed-size chunking (e.g., 500
  characters with overlap) splits semantic units across fragments,
  forcing the generator to reconstruct coherent knowledge from partial
  context.
\item
  \textbf{Static corpora.} Traditional RAG retrieves from a pre-indexed
  document collection. When user queries fall outside the corpus, the
  system fails silently or hallucinates.
\item
  \textbf{Retrieval noise.} Embedding-based similarity search returns
  fragments that are textually similar but not always semantically
  relevant, and the generator must filter signal from noise in a single
  forward pass.
\item
  \textbf{Entangled capabilities.} Existing systems treat retrieval as
  supplementary to a model's parametric knowledge, creating an ambiguous
  boundary between what the model ``knows'' internally and what was
  retrieved externally. This entanglement makes it impossible to audit,
  update, or control the knowledge base independently of the model.
\end{enumerate}

\subsubsection{1.2 A Biological
Perspective}\label{a-biological-perspective}

Biology separates cognitive capability from acquired knowledge: DNA
encodes the architecture of a brain --- structure, connectivity, innate
cognitive machinery --- but contributes zero factual knowledge. All
facts are acquired through experience, stored in neural structures that
are continuously updated, consolidated, and pruned. Evolution optimizes
the cognitive architecture slowly across generations; knowledge
acquisition happens rapidly within a single lifetime.

Biological memory further exhibits a two-phase pattern: during waking
hours, new experiences are rapidly encoded in the hippocampus (a fast
staging buffer); during sleep, this buffer is consolidated into cortical
long-term memory --- redundant memories merged, weak associations
pruned, and the overall structure reorganized.

\subsubsection{1.3 From Biology to
Architecture}\label{from-biology-to-architecture}

Evolve translates these biological principles into a computational
architecture for language model systems:

{\def\LTcaptype{none} 
\begin{longtable}[]{@{}
  >{\raggedright\arraybackslash}p{(\linewidth - 2\tabcolsep) * \real{0.5000}}
  >{\raggedright\arraybackslash}p{(\linewidth - 2\tabcolsep) * \real{0.5000}}@{}}
\toprule\noalign{}
\begin{minipage}[b]{\linewidth}\raggedright
Biological Principle
\end{minipage} & \begin{minipage}[b]{\linewidth}\raggedright
Architectural Analog
\end{minipage} \\
\midrule\noalign{}
\endhead
\bottomrule\noalign{}
\endlastfoot
DNA encodes cognitive architecture, not facts & Model weights encode
reasoning capability; factual knowledge is externalized \\
Experience-based knowledge acquisition & Teacher-compiled knowledge
sections stored in an external knowledge store, with category-routed
expert teachers \\
Hippocampal short-term buffer & Staging store for newly acquired
knowledge \\
Cortical long-term consolidation & Canonical store, the authoritative
knowledge base \\
Sleep consolidation (merge, prune, reorganize) & Offline sleep phase
with teacher-mediated compilation \\
Evolutionary efficiency (minimal architecture, maximal capability) &
Small local model (2B--7B) operating as a pure reasoning engine \\
\end{longtable}
}

The key distinction from standard RAG: conventional retrieval-augmented
systems retrieve per query and forget --- each query starts fresh
against a static corpus. Evolve compiles knowledge that persists,
accumulates, and improves over time. The knowledge store is not a
document index but a living, teacher-compiled knowledge base that grows
with use and is refined through consolidation.

The architecture supports a clean separation of reasoning from
knowledge: the local model's weights encode general reasoning capability
while all factual knowledge resides in the external knowledge store.
Evolve offers two generation modes over this architecture ---
\textbf{suppress} (strict section-only grounding, a zero-knowledge
stance appropriate for queries where auditability matters) and
\textbf{augment} (sections plus parametric fallback, appropriate when
section content quality varies or parametric supplementation of factual
answers is acceptable). Both modes operate over the same knowledge
lifecycle.

\subsubsection{1.4 Deployment Flexibility}\label{deployment-flexibility}

The clean separation of reasoning from knowledge creates a
deployment-agnostic design. On \textbf{edge devices}, teacher calls
amortize rapidly as the knowledge store matures --- the system becomes
largely self-sufficient, with queries and knowledge remaining on-device
after initial acquisition. On \textbf{enterprise systems}, the knowledge
store provides governance and auditability (every claim traces to a
specific section with provenance), cost amortization across users
(teacher costs paid once, reused by all), and independent evolution of
components (model upgrades preserve the accumulated knowledge artifact).

\subsubsection{1.5 Contributions}\label{contributions}

The central contribution of this paper is the \textbf{persistent
knowledge lifecycle} --- an online pipeline (classify → retrieve →
acquire → refresh → generate) together with an offline consolidation
pass, operating over an evolving teacher-compiled knowledge store ---
established as a first-class system primitive for small-model knowledge
access, replacing the stateless retrieve-then-generate pattern of
conventional RAG. The five technical contributions below are the
mechanisms that realize it:

\begin{enumerate}
\def\labelenumi{\arabic{enumi}.}
\item
  \textbf{Teacher-compiled persistent knowledge.} Knowledge enters the
  system as dense, encyclopedic sections compiled by teacher models at
  natural conceptual boundaries --- not as retrieved document chunks.
  These sections are reusable knowledge primitives that persist and
  improve across queries, replacing stateless retrieval with persistent
  knowledge compilation.
\item
  \textbf{Knowledge lifecycle with sleep consolidation.} A two-tier
  memory architecture (staging → canonical) with offline
  teacher-mediated consolidation that deduplicates, merges, and
  reorganizes knowledge. Empirically, consolidation reduces store size
  by 31--33.5\% across three independent benchmarks while preserving
  accuracy and modestly lifting suppress-mode accuracy on the custom
  benchmark (+1pp).
\item
  \textbf{Section-based retrieval units outperform chunking.} A
  pluggable retrieval layer enables direct comparison: teacher-compiled
  sections consistently outperform standard RAG-style chunking by 5--9pp
  across every lifecycle condition, delivering higher accuracy from
  fewer, more coherent context blocks.
\item
  \textbf{Teacher-assigned TTL as a knowledge-freshness primitive.} Each
  section carries an explicit refresh duration set by the teacher at
  acquisition time --- minutes for live prices, years for stable facts.
  Expired sections are refreshed inline on cache hit. This is a
  lifecycle primitive for knowledge currency that conventional stateless
  RAG lacks.
\item
  \textbf{Empirical validation across independent benchmarks.} On 750
  benchmark queries across three distributions (custom knowledge,
  NaturalQuestions, TriviaQA), a 2B-parameter model achieves 40--52pp
  cold-cache accuracy gains over its parametric baseline. Full-lifecycle
  evaluation (cold → warm → post-consolidation) on all three benchmarks,
  dual-mode generation (suppress vs augment), threshold sensitivity at
  0.75 and 0.80, Wilson 95\% confidence intervals on all accuracy
  measurements, and MMLU tradeoff characterization provide a multi-axis
  systems evaluation.
\end{enumerate}

Together, these results demonstrate a shift in the cost--accuracy
frontier of knowledge access for small language models: persistent
knowledge reuse and lifecycle refinement maintain or improve accuracy
across lifecycle stages while reducing teacher invocation by over 50\%,
establishing knowledge persistence as a primary factor in performance.

\begin{center}\rule{0.5\linewidth}{0.5pt}\end{center}

\subsection{2. Related Work}\label{related-work}

\subsubsection{2.1 Retrieval-Augmented
Generation}\label{retrieval-augmented-generation}

RAG was introduced by Lewis et al.~(2020) as a method to combine
parametric and non-parametric memory in language models. Standard RAG
retrieves document chunks and conditions generation on them. While
effective at grounding generation in external knowledge, standard RAG
treats the retrieval corpus as a static, pre-indexed collection and
relies on fixed-size chunking that frequently splits semantic units.
Evolve extends this paradigm by replacing passive chunk retrieval with
active knowledge compilation, replacing static corpora with an
organically growing knowledge store, and offering two generation modes
over a shared knowledge lifecycle: strict section-only grounding or
sections plus parametric fallback.

\subsubsection{2.2 Self-RAG: Adaptive Retrieval
Decisions}\label{self-rag-adaptive-retrieval-decisions}

Asai et al.~(2023) introduced Self-RAG, where a language model learns to
adaptively decide when to retrieve, whether retrieved passages are
relevant, and whether its own output is supported by evidence. The model
generates special reflection tokens --- Retrieve, IsRelevant,
IsSupported, and IsUseful --- as part of next-token prediction, enabling
per-segment retrieval decisions without a separate critic model.
Self-RAG (7B and 13B parameters) significantly outperformed ChatGPT and
retrieval-augmented Llama2-chat across open-domain QA, reasoning, and
fact verification tasks.

Self-RAG addresses adaptive retrieval within a single generation turn by
training a model to emit reflection tokens. Evolve takes a different
approach: rather than asking the model to decide whether each generated
segment needs retrieval, it commits to a single per-query retrieval pass
and externalizes all factual knowledge into a persistent, evolving store
that survives across sessions. The two systems target different problems
--- Self-RAG optimizes within-turn retrieval timing, Evolve optimizes
long-term knowledge accumulation.

\subsubsection{2.3 RAPTOR: Hierarchical Knowledge
Organization}\label{raptor-hierarchical-knowledge-organization}

Sarthi et al.~(2024) proposed RAPTOR, which constructs a hierarchical
tree of summaries from the bottom up: raw text chunks are embedded,
clustered using Gaussian Mixture Models, and summarized recursively
until a complete tree is formed. At query time, the system retrieves
from any level of the tree, accessing both fine-grained details and
high-level thematic summaries. Coupled with GPT-4, RAPTOR improved the
best-known performance on the QuALITY benchmark by 20\% in absolute
accuracy.

RAPTOR and Evolve both move beyond flat chunk retrieval, but in
fundamentally different ways. RAPTOR builds a hierarchical tree of
algorithmically generated summaries from a fixed corpus --- a static
retrieval optimization. Evolve uses a flat, category-scoped store of
teacher-compiled semantic sections --- a living structure that acquires
new knowledge through teacher consultation, consolidates through sleep
cycles, and refreshes through TTL-driven inline updates. RAPTOR's tree
is built once and never updated; Evolve's store grows and refines with
every query. RAPTOR optimizes retrieval; Evolve manages a knowledge
lifecycle.

\subsubsection{2.4 FLARE: Confidence-Based Active
Retrieval}\label{flare-confidence-based-active-retrieval}

Jiang et al.~(2023) introduced FLARE (Forward-Looking Active REtrieval),
which enables a model to iteratively retrieve information throughout
generation, triggered by its own uncertainty. The model generates a
temporary next sentence, checks token-level confidence, and if any
tokens fall below a threshold, uses the low-confidence sentence as a
retrieval query. This forward-looking approach means retrieval targets
what the model actually needs next, rather than what it has already
said.

FLARE introduces confidence-based retrieval triggering during
generation: a model retrieves new content whenever its next-token
confidence drops. FLARE operates at the sentence level during a single
generation pass and has no concept of knowledge persistence, structured
storage, or adaptive refresh intervals. Evolve takes the opposite
approach --- it commits to retrieval up front, before generation begins,
and trusts the local model to compose from the assembled section set in
a single forward pass. The two systems address different sub-problems:
FLARE addresses \emph{when} to retrieve mid-generation, while Evolve
addresses \emph{what} knowledge to accumulate across sessions.

\subsubsection{2.5 MemWalker: Tree-Structured
Navigation}\label{memwalker-tree-structured-navigation}

Chen et al.~(2023) proposed MemWalker, which treats the LLM as an
interactive agent navigating a tree-structured memory. A long document
is segmented and summarized recursively into a tree. At query time, the
LLM navigates this tree top-down through multiple generation steps ---
at each level, it sees only that level's summaries, picks the most
relevant branch, and descends. A working memory accumulates context
across steps until a leaf segment is reached and the final answer is
generated.

MemWalker and Evolve both go beyond flat retrieval, but through
different mechanisms. MemWalker uses hierarchical tree navigation
requiring large models (70B+) for effective chain-of-thought traversal.
Evolve uses flat, category-scoped vector search --- no tree, no
navigation, no chain-of-thought --- enabling effective operation with
small models (2B--7B). MemWalker's tree is built from a single static
document and is never updated, pruned, or refreshed. Evolve's store
acquires knowledge over time, consolidates through sleep cycles, and
refreshes through TTL-driven updates.

\subsubsection{2.6 Adaptive Retrieval Timing and Knowledge
Freshness}\label{adaptive-retrieval-timing-and-knowledge-freshness}

Growing attention has been directed toward when to retrieve and when
retrieved knowledge has become stale. Self-knowledge assessment methods
like SKR train classifiers to predict retrieval need per query.
Time-aware methods like TA-ARE add temporal classifiers for
time-sensitive queries. Confidence-based methods (FLARE, DRAGIN) trigger
retrieval on generation uncertainty.

These approaches address pieces of knowledge freshness management but in
fundamentally simpler ways: binary triggers (retrieve or don't), static
classifiers, or simple confidence thresholds. Evolve's per-section TTL
with usage-driven inline refresh --- where the teacher itself sets an
explicit refresh duration on each acquisition based on how volatile the
underlying knowledge is --- represents a more targeted approach to
knowledge lifecycle management.

\subsubsection{2.7 Sleep-Inspired Memory
Consolidation}\label{sleep-inspired-memory-consolidation}

Multiple research efforts have translated biological sleep consolidation
into neural network training to combat catastrophic forgetting. Sleep
Replay Consolidation (SRC; Tadros et al., 2022) mimics REM sleep-like
activity by converting trained networks to spiking neural networks and
applying unsupervised Hebbian learning during an offline phase.
McClelland et al.~(2022) modeled hippocampal-neocortical interactions
during simulated NREM/REM sleep cycles, where NREM features tightly
coupled replay for recent memories and REM allows freer exploration of
existing knowledge. NeuroDream (Tutuncuoglu, 2024) proposes an explicit
``dream phase'' with internally generated simulations based on stored
latent embeddings.

The sleep consolidation literature provides strong biological and
computational justification for Evolve's scheduled consolidation
process. The parallels are clear: Evolve's staging-to-canonical
reconciliation, with deduplication, merging, and resegmentation,
directly mirrors the NREM/REM sleep cycle's role in strengthening,
pruning, and reorganizing memories. However, existing work operates at
the synaptic weight level within neural networks, using replay to adjust
model parameters. Evolve operates at the knowledge representation level
--- changing the external knowledge available to an unchanged base
model. This is a fundamentally different layer of abstraction: rather
than modifying model weights through sleep-inspired replay, the system
applies sleep-inspired consolidation to an external, structured
knowledge store while the reasoning model remains frozen.

\subsubsection{2.8 Knowledge Distillation}\label{knowledge-distillation}

The teacher-student paradigm in Evolve parallels knowledge distillation
(Hinton et al., 2015), but instead of distilling model weights, Evolve
distills explicit knowledge artifacts --- structured, encyclopedic
sections that persist and grow independently of either model. The
knowledge store is a first-class data structure, not a compressed
approximation of the teacher's parameters.

\subsubsection{2.9 Memory-Augmented and Graph-Based
Retrieval}\label{memory-augmented-and-graph-based-retrieval}

Recent work has explored persistent memory and graph-based approaches
that, like Evolve, go beyond stateless document retrieval. HippoRAG
(Gutiérrez et al., 2024) models retrieval as a hippocampal indexing
process, constructing a knowledge graph from passages and using
Personalized PageRank for multi-hop retrieval. MemoRAG (Qian et al.,
2024) introduces a dual-system architecture with a lightweight model for
global memory formation and a more expressive model for retrieval-guided
generation. Memory3 (Yang et al., 2024) proposes explicit memory as a
third component alongside model weights and context, storing knowledge
in compressible memory tokens.

Evolve differs from these approaches in four ways. First, knowledge is
\textbf{written by a teacher LLM}, not extracted from documents or
compressed from model state. The teacher produces complete, topically
coherent sections --- a quality of knowledge representation that
algorithmic extraction cannot match. Second, all knowledge
reconciliation is \textbf{teacher-mediated}: when sections overlap, the
teacher reads both and produces a coherent compilation, rather than
relying on algorithmic graph merging or memory compression. Third, each
section carries a \textbf{teacher-assigned TTL} reflecting knowledge
volatility, enabling usage-driven refresh that none of these systems
support. Fourth, the architecture supports \textbf{two generation modes}
over the same lifecycle --- strict section-only grounding (suppress) and
sections-plus-parametric-fallback (augment) --- allowing deployment-time
choice based on task requirements and auditability needs.

\subsubsection{2.10 Positioning}\label{positioning}

Existing work addresses individual aspects of the problem space:
Self-RAG addresses retrieval timing, RAPTOR addresses hierarchical
organization, FLARE addresses confidence-based retrieval, MemWalker
addresses tree navigation, HippoRAG addresses graph-based hippocampal
retrieval, MemoRAG addresses dual-system memory, and sleep consolidation
research addresses offline memory reorganization. Rather than extending
any of these approaches, Evolve introduces a unified knowledge lifecycle
system --- teacher-compiled knowledge, category-aware classification,
section-based retrieval, two-tier memory with sleep consolidation,
usage-driven refresh, and dual-mode generation --- where each component
serves the central goal of persistent, evolving knowledge that improves
with use.

\begin{center}\rule{0.5\linewidth}{0.5pt}\end{center}

\subsection{3. Architecture Overview}\label{architecture-overview}

Evolve is a system where queries populate and refine a persistent
knowledge base over time, rather than retrieving from a fixed corpus per
query. At a high level, it operates through six lifecycle phases:

\begin{enumerate}
\def\labelenumi{\arabic{enumi}.}
\tightlist
\item
  \textbf{Classify} --- the local model determines what knowledge is
  needed and which domain categories to search, producing a query type
  label and targeted search texts for each category.
\item
  \textbf{Retrieve} --- the search texts are embedded and used to search
  both the staging and canonical vector collections within each
  category, gathering candidate sections.
\item
  \textbf{Acquire} --- on cache miss, the category-appropriate teacher
  compiles a new knowledge section; on cache hit, stored sections are
  used directly.
\item
  \textbf{Refresh} --- on cache hit of expired sections, the teacher is
  invoked inline to update them with current information.
\item
  \textbf{Generate} --- the local model produces a grounded response
  from the assembled section set in the configured generation mode.
\item
  \textbf{Consolidate} --- an offline sleep cycle deduplicates, merges,
  and reorganizes the staging store into a refined canonical store
  through teacher-mediated compilation.
\end{enumerate}

These phases are supported by five principal components: a
\textbf{teacher router} (dispatching to one or more configurable teacher
models), a \textbf{local model} (the reasoning engine that handles
classification and generation), an \textbf{embedding model}, a
\textbf{pluggable retrieval store} (with staging and canonical
collections), and a \textbf{metadata store}. An \textbf{orchestrator}
coordinates the online query pipeline (classify through generate), while
consolidation runs as a separate offline process.

\begin{figure}
\centering
\pandocbounded{\includegraphics[keepaspectratio,alt={Evolve Architecture}]{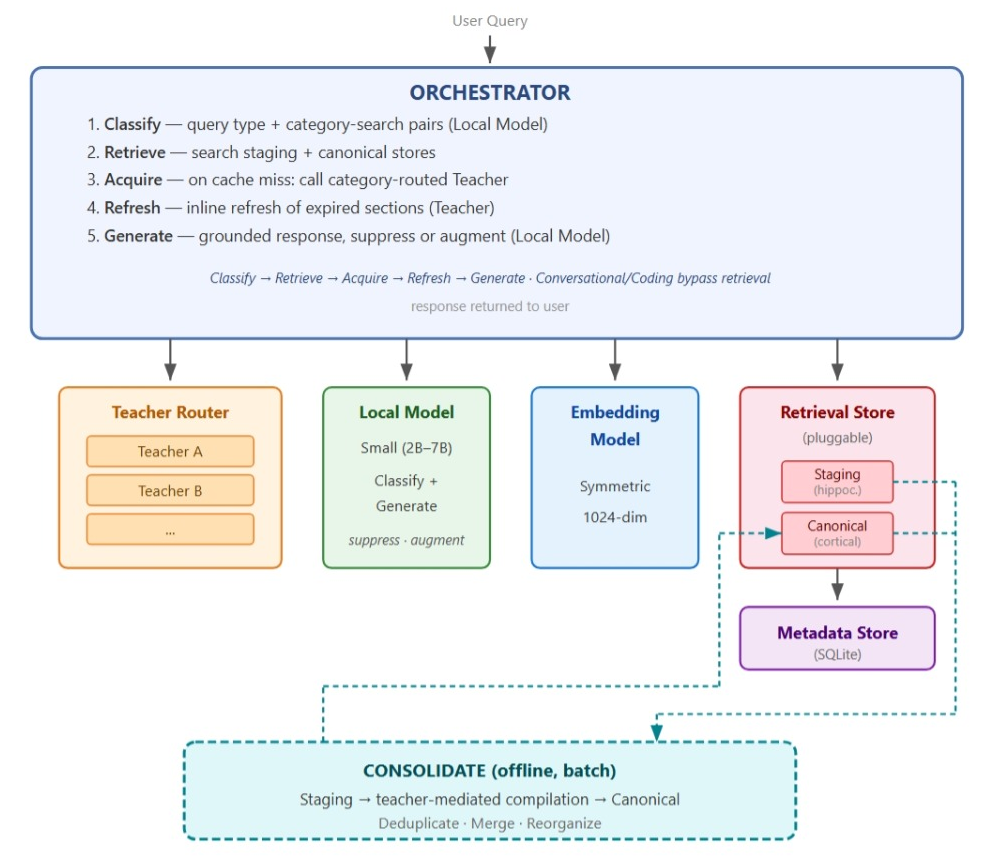}}
\caption{Evolve Architecture}
\end{figure}

\textbf{Figure 1.} Evolve architecture. The local model is the reasoning
engine --- handling query classification and grounded answer generation
--- while all factual knowledge resides in the external knowledge store.
The teacher router dispatches knowledge compilation requests to
category-appropriate teacher models --- invoked only during knowledge
acquisition, refresh, and compilation, never to answer user queries
directly. The staging store (hippocampal analog) receives all new
knowledge before reconciliation; the canonical store (cortical analog)
holds consolidated, authoritative knowledge. Sleep consolidation is the
sole gate between them.

\subsubsection{3.1 The Section: Atomic Knowledge
Unit}\label{the-section-atomic-knowledge-unit}

The fundamental data unit is the \textbf{section} --- a self-contained,
encyclopedic article on a single semantic topic. Unlike document chunks
produced by character-count or paragraph-based splitting, each section
is produced by the teacher as a single, dense knowledge module at
natural conceptual boundaries. The teacher is explicitly prompted to
maximize factual density --- including specific numerical data (dates,
measurements, quantities), named entities (people, places,
institutions), and verifiable facts --- producing content at the level
of a Wikipedia Featured Article rather than a surface-level summary.
Each section is:

\begin{itemize}
\tightlist
\item
  \textbf{Semantically coherent}: Covers exactly one topic boundary
  determined by the teacher's understanding of the domain, not by
  character count or paragraph breaks.
\item
  \textbf{Self-contained}: Readable and useful without reference to
  other sections, analogous to a Wikipedia article.
\item
  \textbf{Reusable}: Not tailored to the query that triggered its
  creation. A general-purpose knowledge module that serves any future
  query touching the same topic.
\end{itemize}

Each section carries metadata:

{\def\LTcaptype{none} 
\begin{longtable}[]{@{}
  >{\raggedright\arraybackslash}p{(\linewidth - 2\tabcolsep) * \real{0.3500}}
  >{\raggedright\arraybackslash}p{(\linewidth - 2\tabcolsep) * \real{0.6500}}@{}}
\toprule\noalign{}
\begin{minipage}[b]{\linewidth}\raggedright
Field
\end{minipage} & \begin{minipage}[b]{\linewidth}\raggedright
Description
\end{minipage} \\
\midrule\noalign{}
\endhead
\bottomrule\noalign{}
\endlastfoot
\texttt{id} & Unique identifier (UUID) \\
\texttt{topic} & Semantic topic name (e.g., ``Rayleigh Scattering'') \\
\texttt{summary} & One-sentence summary of the content \\
\texttt{content} & Full encyclopedic text \\
\texttt{refresh\_minutes} & TTL in minutes (normalized from the
teacher-specified duration) \\
\texttt{category} & Domain category from a managed taxonomy \\
\texttt{created\_at} & Creation timestamp \\
\texttt{store} & Current location: staging or canonical \\
\end{longtable}
}

This is a critical distinction from standard RAG: the teacher produces
reusable, topically segmented knowledge rather than a one-time tailored
response. The knowledge store is populated with durable modules, not
ephemeral answers. Sections are stable retrieval units at any given
lifecycle stage, but may be structurally transformed across stages
through teacher-mediated refresh (Section 4.4) and consolidation
(Section 5) --- a split, merge, or compilation produces a new section
set that replaces the inputs.

\subsubsection{3.2 Teacher Router and Teacher
Models}\label{teacher-router-and-teacher-models}

Knowledge compilation is handled by one or more large, capable teacher
LLMs (e.g., GPT-4o, Claude, Llama 70B), each used exclusively for
knowledge compilation --- never to answer user queries directly. A
\textbf{teacher router} manages the mapping between domain categories
and teacher models, dispatching each compilation request to the
appropriate teacher based on the category of the knowledge being
acquired, refreshed, or compiled.

\paragraph{Category-Level Teacher
Assignment}\label{category-level-teacher-assignment}

Each teacher model is registered with an optional list of domain
categories it is responsible for. When the orchestrator needs to
acquire, refresh, or compile knowledge in a given category, the router
selects the teacher assigned to that category. If no category-specific
teacher is configured, the router falls back to a configurable default
teacher.

This category-level routing enables several deployment patterns:

\begin{itemize}
\tightlist
\item
  \textbf{Expert teachers}: A medical LLM (e.g., a fine-tuned clinical
  model) handles categories like Medicine and Biology, while a
  general-purpose teacher handles the remaining domains. Each teacher
  operates within its area of strongest competence.
\item
  \textbf{Enterprise integration}: Organizations can route proprietary
  domain categories to internal models trained on corporate knowledge,
  while general academic categories use a cloud-hosted teacher.
\item
  \textbf{Cost optimization}: Expensive frontier models can be reserved
  for categories requiring the highest accuracy (e.g., Law, Medicine),
  while less expensive models handle categories where quality
  requirements are lower.
\item
  \textbf{Regulatory compliance}: Categories subject to regulatory
  requirements (e.g., financial or healthcare knowledge) can be routed
  to approved, auditable models, while other categories use standard
  teachers.
\end{itemize}

The router is transparent to the rest of the pipeline --- the
orchestrator requests knowledge compilation for a category, and the
router handles teacher selection. Adding, removing, or reassigning
teachers requires only configuration changes, not code modifications.

\paragraph{Teacher Operations}\label{teacher-operations}

Each teacher performs three operations:

\begin{enumerate}
\def\labelenumi{\arabic{enumi}.}
\item
  \textbf{Acquire}: Given a query and a domain category, produce a
  single dense, fact-packed encyclopedic section covering the topic. The
  teacher generates a reusable knowledge module, not a query-specific
  answer. Each teacher call produces exactly one section; the
  classifier's per-pair architecture handles breadth by issuing separate
  teacher calls for each domain.
\item
  \textbf{Refresh}: Given expired sections, produce updated versions
  with current information. The teacher may restructure, split, or merge
  topics as warranted by changes in the knowledge landscape. Critically,
  the teacher may also adjust the refresh duration --- changing a
  section from 1 month to 1 week if the topic has become more volatile.
  Each section carries an independent lifecycle.
\item
  \textbf{Compile}: Given overlapping sections from the staging and
  canonical stores, produce a single authoritative, deduplicated set.
  The structural outcome is implicit in the teacher's output: if one
  section went in and two came out, a split occurred; if two went in and
  one came out, a merge occurred. The system needs no explicit
  structural classification.
\end{enumerate}

The acquire operation returns a structured JSON format with a single
section:

\begin{Shaded}
\begin{Highlighting}[]
\FunctionTok{\{}
  \DataTypeTok{"query\_context"}\FunctionTok{:} \StringTok{"original query or context that triggered this call"}\FunctionTok{,}
  \DataTypeTok{"section"}\FunctionTok{:} \FunctionTok{\{}
    \DataTypeTok{"topic"}\FunctionTok{:} \StringTok{"Specific semantic topic name"}\FunctionTok{,}
    \DataTypeTok{"refresh"}\FunctionTok{:} \FunctionTok{\{}\DataTypeTok{"value"}\FunctionTok{:} \DecValTok{1}\FunctionTok{,} \DataTypeTok{"unit"}\FunctionTok{:} \StringTok{"year"}\FunctionTok{\},}
    \DataTypeTok{"summary"}\FunctionTok{:} \StringTok{"One{-}sentence summary"}\FunctionTok{,}
    \DataTypeTok{"content"}\FunctionTok{:} \StringTok{"Full encyclopedic treatment..."}
  \FunctionTok{\}}
\FunctionTok{\}}
\end{Highlighting}
\end{Shaded}

\subsubsection{3.3 Local Model: The Reasoning
Engine}\label{local-model-the-reasoning-engine}

The local model is a small, inexpensive LLM (2B--7B parameters) that
handles all real-time user-facing operations --- classification and
generation. It contributes only linguistic competence (grammar,
coherence, fluency) and logical reasoning (inference, comparison,
synthesis) --- capabilities encoded in its weights during training,
analogous to the innate cognitive machinery encoded in DNA, while
factual knowledge is externalized to the knowledge store. The
architecture supports two generation modes --- suppress and augment ---
over the same lifecycle; one is selected at deployment time for all
factual queries that flow through the retrieval pipeline:

\begin{itemize}
\item
  \textbf{Suppress mode} (strict grounding / zero-knowledge stance). The
  system prompt instructs the model to suppress parametric knowledge,
  the generate prompt requires every claim to be backed by a referenced
  section, and the pipeline ensures the model never receives a factual
  query without accompanying sections. Any claim not supported by a
  section is treated as incorrect, regardless of whether parametric
  knowledge happens to contain the correct answer. This is the
  appropriate mode when auditability, provenance, or consistent
  grounding matter --- regulated domains, citation-requiring
  applications, or any deployment that must trace every factual claim to
  a specific source.
\item
  \textbf{Augment mode} (flexible grounding). The system prompt permits
  the model to supplement retrieved sections with parametric knowledge
  where helpful. Sections remain the primary source, but the model can
  fill gaps when section coverage is partial or when the model's own
  knowledge is material to the task. This is the appropriate mode when
  section content quality varies across queries or when parametric
  supplementation can materially improve factual answers without
  violating auditability requirements.
\end{itemize}

The generation-mode policy is a deployment-level configuration with two
settings: \texttt{suppress} and \texttt{augment}. Each applies the
corresponding mode uniformly to every factual query that flows through
the retrieval pipeline. Both configurations are reported in Section 7
(dual-mode evaluation).

Two query-type categories bypass the retrieval pipeline entirely via
classifier routing, independent of the generation-mode setting:
\texttt{conversational} queries (greetings, casual chat, arithmetic,
opinions) go directly to the local model without retrieval, and
\texttt{coding} queries also bypass. The coding bypass is an empirical
design choice --- teacher-compiled sections for coding questions tend to
be descriptive (architecture overviews, API descriptions) rather than
exemplary (working code), and constraining the local model to generate
from those sections materially degrades code output quality. A dedicated
code-generation mode that treats retrieved sections as hints for a
code-specialized generation prompt is future work (Section 8.1).

\paragraph{Why Not Self-Assess Knowledge
Boundaries?}\label{why-not-self-assess-knowledge-boundaries}

An alternative design would allow the local model to first attempt
answering from parametric memory and only consult the knowledge store
when it determines its own knowledge is insufficient. We rejected this
approach because small models are unreliable judges of their own
knowledge boundaries. On a 250-question stratified subset of the MMLU
benchmark, the 2B local model answers 28\% of questions incorrectly with
full confidence --- it does not refuse or hedge, it confidently states
the wrong answer. A model that cannot distinguish what it knows from
what it doesn't know cannot be trusted to decide when external knowledge
is needed. Metacognitive self-assessment is itself a capability that
scales with model size; research on MemWalker found that reliable
self-navigation required 70B+ parameter models.

The architecture sidesteps this unreliable self-assessment by always
grounding responses in externalized knowledge through the retrieval
pipeline --- both suppress and augment modes depend on retrieval, they
differ only in whether the generation step is permitted to supplement
with parametric knowledge. Suppress mode's strict stance produces a
small accuracy cost on well-rehearsed material (Section 7) but ensures
consistent, auditable grounding across all queries --- a property
essential for deployment in domains where hallucination carries real
consequences.

The local model performs two operations:

\begin{enumerate}
\def\labelenumi{\arabic{enumi}.}
\item
  \textbf{Classify}: Given a query and a domain taxonomy, produce (a) a
  \textbf{query type} label (\texttt{factual}, \texttt{coding}, or
  \texttt{conversational}) characterizing the nature of the query, and
  (b) a list of \textbf{category-search pairs} --- each pairing a
  specific domain category with a focused search text tailored for
  vector retrieval within that domain. A cross-domain query such as
  ``the electrochemistry of lithium-ion battery recycling'' is labelled
  \texttt{factual} and produces pairs for both Chemistry and
  Engineering, each with a domain-specific search text optimized for
  retrieval in that category. The type label determines routing ---
  \texttt{conversational} and \texttt{coding} queries bypass the
  retrieval pipeline entirely; only \texttt{factual} queries continue
  into retrieval and generation. The category-search pairs drive
  retrieval for the factual path.
\item
  \textbf{Generate}: Given a query and sections, produce a response
  grounded exclusively in the provided sections, maintaining an expert
  voice with no meta-references to sources.
\end{enumerate}

\subsubsection{3.4 Two-Tier Memory
Architecture}\label{two-tier-memory-architecture}

Inspired by the hippocampal-cortical memory system, embeddings are
organized into two collections:

\begin{itemize}
\tightlist
\item
  \textbf{Staging store} (hippocampal analog): Where all new knowledge
  lands --- from new acquisition or inline refresh. Actively queried
  during operation. Contents have not yet been reconciled with the
  canonical store. Accepts temporary redundancy in exchange for
  immediate availability.
\item
  \textbf{Canonical store} (cortical analog): The consolidated,
  deduplicated, authoritative knowledge base. Only modified by the sleep
  consolidation cycle, which is the sole gate from staging to canonical.
\end{itemize}

This design ensures a single entry point for all new knowledge (the
staging store) and a single reconciliation path (sleep consolidation).
Search queries both collections simultaneously, with results merged and
ranked by cosine similarity. Because the classifier produces multiple
category-search pairs per query, retrieval spans all selected domains
--- each pair searches independently within its category, and results
are merged across pairs. This per-pair design ensures that
interdisciplinary queries retrieve knowledge from every relevant domain
using domain-specific search vocabulary rather than being siloed into a
single category or searched with a one-size-fits-all query. A
configurable similarity threshold eliminates low-relevance matches. The
threshold must be calibrated to the embedding model's discrimination
characteristics: too low admits related-but-insufficient sections from
the same broad topic, causing the system to serve cached content that
doesn't address the specific question; too high causes legitimate cache
misses on queries that differ slightly in phrasing from the original
acquisition query. Empirical testing with mxbai-embed-large-v1 (a
1024-dimensional symmetric embedding model) found that a threshold of
0.75--0.80 provides the best balance between retrieval precision and
cache reuse.

\subsubsection{3.5 Metadata Store}\label{metadata-store}

A relational database (SQLite) stores full section content and lifecycle
metadata. The vector store holds only embeddings and lightweight
metadata (topic, summary, categories); the metadata store is the source
of truth for section content, TTL tracking, and store membership.

\subsubsection{3.6 Domain Taxonomy and Category
Normalization}\label{domain-taxonomy-and-category-normalization}

The domain taxonomy defines the category vocabulary used by the
classifier. The classifier produces category-search pairs whose category
values must match entries in this taxonomy exactly.

\paragraph{Flat Taxonomy Design}\label{flat-taxonomy-design}

The taxonomy is a single flat list of 42 categories spanning the full
breadth of human knowledge --- from academic disciplines (Physics,
Chemistry, Biology, Mathematics) through professional domains (Law,
Finance, Engineering) to everyday life topics (Cooking and Food, Home
and DIY, Fitness and Sports, Parenting and Family). The list is
maintained as a standalone JSON file, loaded at startup, and injected
into every prompt that produces category-search pairs.

A flat taxonomy was chosen over hierarchical grouping after empirical
testing revealed that two-level structures (e.g., ``Sciences → Physics,
Chemistry, Biology'') introduce more problems than they solve:

\begin{itemize}
\tightlist
\item
  \textbf{No token savings.} A two-level taxonomy still requires listing
  every subcategory in the prompt, since the LLM must see the full
  vocabulary to select from it. The group labels add tokens without
  reducing the leaf set.
\item
  \textbf{Forced artificial boundaries.} Many categories sit naturally
  at the intersection of multiple parent groups. Psychology could belong
  under Medicine, Social Sciences, or Cognitive Science depending on the
  query. A hierarchical structure forces the LLM to navigate a tree
  path, introducing routing ambiguity at each branch point. A flat list
  lets the LLM match directly to the most appropriate category without
  resolving which parent group it ``belongs'' to.
\item
  \textbf{Small model reliability.} Small models (2B--7B) are more
  reliable at selecting from a flat list than at navigating a
  hierarchical classification tree, where errors at higher levels
  cascade into incorrect subcategory assignments.
\end{itemize}

The 42-category vocabulary balances granularity against routing
ambiguity: categories are specific enough to provide meaningful search
gating (separating Physics from Chemistry, Statistics from Mathematics)
without being so fine-grained that the LLM struggles to distinguish
between them.

\paragraph{Category Normalization}\label{category-normalization}

To prevent mismatches caused by case variation or minor naming
deviations in LLM output, all category values are normalized at every
boundary where they enter or leave the system:

\begin{itemize}
\tightlist
\item
  \textbf{LLM output normalization.} Every list of category-search pairs
  produced by the local model passes through a normalization step that
  maps each category string to its canonical form via case-insensitive
  lookup with substring fallback. Pairs with unrecognizable categories
  are dropped with a warning.
\item
  \textbf{Storage normalization.} A dedicated storage gateway normalizes
  categories on every section write --- whether from teacher acquisition
  or inline refresh --- ensuring that the vector store and metadata
  store contain only canonical category names.
\item
  \textbf{Query-time matching.} Because all stored categories are
  guaranteed canonical, vector store search uses exact string matching
  against query categories that have already been normalized at the LLM
  output boundary. This eliminates the runtime cost of case-insensitive
  comparison during retrieval.
\end{itemize}

This normalization chain ensures that a section stored under ``Computer
Science'' is always found when the classifier outputs ``computer
science'' or ``Computer science'', without relying on case-insensitive
comparison at search time.

\begin{center}\rule{0.5\linewidth}{0.5pt}\end{center}

\subsection{4. Query Pipeline}\label{query-pipeline}

The orchestrator executes a multi-step pipeline for each user query,
with conditional branches that adapt behavior based on cache state and
content freshness.

\subsubsection{4.1 Query
Contextualization}\label{query-contextualization}

In interactive mode, the system maintains the full conversation history
as user/assistant turn pairs. All turns are preserved --- the model's
context window is the natural limit, matching standard chat behavior. On
each new query, the conversation history is prepended as context so the
classifier can resolve anaphoric references (e.g., ``what about its
melting point?'' after discussing cesium) and produce appropriate
domain-specific search pairs. Only the natural-language response is
stored in the history --- no retrieved sections, no internal pipeline
state, no section references. The knowledge store handles factual
memory; the conversation history handles only conversational continuity.

\subsubsection{4.2 Category-Based Search
Gating}\label{category-based-search-gating}

The classifier serves a triple role: it assigns the query a type label
(\texttt{factual}, \texttt{coding}, or \texttt{conversational}),
identifies which domain categories contain relevant knowledge, and ---
via the type label --- determines whether retrieval is invoked at all.
Queries classified as \texttt{conversational} (greetings, casual chat,
arithmetic, opinions) or \texttt{coding} bypass the retrieval pipeline
entirely, letting the local model respond directly; only
\texttt{factual} queries flow through the full pipeline, where the
deployment's generation-mode policy (Section 3.3) selects the system
prompt. This ensures the knowledge acquisition pipeline is invoked only
when factual grounding is the right tool --- avoiding unnecessary
teacher calls on conversational queries and avoiding the
descriptive-section-anchoring effect on coding queries (Section 8.1).

\paragraph{Per-Pair Architecture}\label{per-pair-architecture}

Each category-search pair couples a specific domain category with a
search text optimized for retrieval within that domain:

\begin{Shaded}
\begin{Highlighting}[]
\OtherTok{[}
  \FunctionTok{\{}\DataTypeTok{"category"}\FunctionTok{:} \StringTok{"Physics"}\FunctionTok{,} \DataTypeTok{"search"}\FunctionTok{:} \StringTok{"thermodynamics of protein folding"}\FunctionTok{\}}\OtherTok{,}
  \FunctionTok{\{}\DataTypeTok{"category"}\FunctionTok{:} \StringTok{"Biology"}\FunctionTok{,} \DataTypeTok{"search"}\FunctionTok{:} \StringTok{"protein folding mechanisms and stability"}\FunctionTok{\}}
\OtherTok{]}
\end{Highlighting}
\end{Shaded}

The per-pair design is motivated by a key observation: the optimal
search text for Physics (``thermodynamics of protein folding'') differs
from the optimal search text for Biology (``protein folding mechanisms
and stability''). A single shared search text forces every category to
be searched with identical terminology, missing domain-specific
vocabulary that would surface the most relevant sections. By coupling
each category with its own search text, the classifier enables
domain-specific retrieval vocabulary that maximizes the probability of
finding relevant sections in each category's embedding space.

The multi-category design remains critical for interdisciplinary
queries. A question about ``machine learning for drug discovery'' would
produce pairs for both Computer Science and Medicine, each with
domain-appropriate search text. Single-category classification would
force the system to choose one domain and miss relevant knowledge in the
other --- a particularly damaging failure mode given that the most
challenging questions tend to be interdisciplinary.

\paragraph{Search Text Quality
Control}\label{search-text-quality-control}

Empirical testing revealed that LLM-generated search text is susceptible
to several forms of pollution that degrade retrieval quality:

\begin{itemize}
\tightlist
\item
  \textbf{Meta-instruction leakage.} The search text describes an
  answering process rather than a knowledge domain (e.g., ``provide the
  final answer statement'' instead of a subject-matter concept).
\item
  \textbf{Question framing artifacts.} Interrogative phrasing from the
  original query leaks into the search text (e.g., ``what is the
  mechanism of'' instead of the mechanism name itself).
\item
  \textbf{Problem-specific values.} Concrete numbers, measurements, or
  scenario descriptions from the query appear in the search text, making
  it too specific to retrieve general knowledge useful for any problem
  on the same topic.
\end{itemize}

The classifier prompt enforces a \textbf{textbook index heuristic} via
an include mechanism (see Section 6.4): every search text must be a
general knowledge concept that could appear as a chapter heading or
index entry in a textbook. It must never be a restatement of the
question, and never describe an answering process.

The shared rules mandate five stripping operations before any search
text is emitted: (1) strip question framing, (2) strip answer/format
meta-instructions, (3) strip answer choices, (4) generalize
problem-specific values to underlying concepts, and (5) strip
conversational scaffolding. A self-check requires the LLM to verify each
search text against three criteria before output: it could appear as a
textbook entry, it contains zero problem-specific values, and it is
purely about subject-matter knowledge.

\paragraph{Design Considerations}\label{design-considerations}

\begin{itemize}
\tightlist
\item
  \textbf{Speed}: Classification runs on the small local model for every
  query. Producing category-search pairs is essentially a classification
  and keyword extraction task, which small models handle well.
\item
  \textbf{Error asymmetry}: A false positive (an irrelevant
  category-search pair) is cheap --- you search an extra category and
  find nothing useful. A false negative (missing a relevant pair) means
  you miss knowledge you actually have. The system biases toward
  inclusion: when in doubt, the model produces additional pairs rather
  than choosing a single best match.
\item
  \textbf{Deduplication}: The shared instruction fragment prohibits
  duplicate or near-duplicate search concepts across pairs. Two search
  texts that would retrieve the same encyclopedia article are considered
  duplicates --- only one is retained.
\end{itemize}

\subsubsection{4.3 Retrieval and Global
Ranking}\label{retrieval-and-global-ranking}

Each category-search pair is processed independently: the search text is
embedded and used to search both the staging and canonical vector
collections, filtered to the pair's category. Both stores must be
searched because refreshed and newly acquired sections live in the
staging store until the next sleep consolidation cycle.

\textbf{Cache hit}: If results exceed the similarity threshold, they are
collected into a \textbf{store pool} --- a global set of candidate
sections tracked with their similarity scores.

\textbf{Cache miss}: If no results meet the threshold, the
category-appropriate teacher is called via the teacher router to acquire
a new section for that pair (each acquire call produces exactly one
section). The section is embedded, stored in the staging collection and
metadata store, and placed into a separate \textbf{teacher pool} ---
sections that are always included in downstream processing without
competing for ranked positions.

\paragraph{Two-Pool Architecture}\label{two-pool-architecture}

The retrieval stage maintains two distinct pools throughout the query
lifecycle:

\begin{enumerate}
\def\labelenumi{\arabic{enumi}.}
\item
  \textbf{Store pool}: All sections retrieved from the vector store,
  tracked by section ID and best similarity score. After all
  category-search pairs are processed, the store pool is globally ranked
  by similarity and capped at a configurable maximum (default 15). When
  multiple category searches return the same section, the highest
  similarity score is retained. This global rank-and-cap ensures that
  only the most relevant sections across all categories survive,
  regardless of how many category-search pairs the classifier produces.
  Without this cap, a query classified into four categories could
  accumulate 40+ sections, overwhelming the local model with noise.
\item
  \textbf{Teacher pool}: All sections generated by teacher calls (cache
  misses). These are always included in full --- they are freshly
  generated, targeted knowledge produced specifically for the current
  query's needs. They do not compete for ranked positions in the store
  pool.
\end{enumerate}

The combined section set --- capped store pool plus full teacher pool
--- is passed directly to the generation step.

\subsubsection{4.4 Usage-Driven Inline
Refresh}\label{usage-driven-inline-refresh}

On a cache hit, each retrieved section's freshness is validated against
its teacher-assigned TTL. The teacher specifies an explicit duration on
each acquisition --- for example,
\texttt{\{"value":\ 30,\ "unit":\ "minutes"\}} for rapidly changing
data, \texttt{\{"value":\ 6,\ "unit":\ "months"\}} for evolving policy,
or \texttt{\{"value":\ 1,\ "unit":\ "year"\}} for stable facts. The
teacher chooses the duration based on its assessment of how volatile the
underlying knowledge is. The duration is normalized to minutes at parse
time. A section is expired when the elapsed time since creation exceeds
its TTL. Ephemeral sections (TTL = 0) are persisted on acquisition
rather than discarded so that multiple queries within the same acquiring
turn can reuse the section without a redundant teacher call; they are
treated as immediately expired on any subsequent cache hit (triggering
refresh rather than direct serving), and the next sleep consolidation
removes them without compilation.

Critically, refresh is triggered \textbf{inline during query
processing}, not as a batch operation during sleep consolidation. When
expired sections are encountered, they are sent to the teacher for
refresh. The teacher returns updated sections, which are stored in the
staging store, while the expired originals are removed from whichever
store they came from (staging or canonical). A canonical section that
triggered refresh is therefore demoted: its successor re-enters through
staging and is re-evaluated by the next sleep consolidation cycle.

This usage-driven design has several important properties:

\begin{itemize}
\tightlist
\item
  \textbf{Only actively queried knowledge is refreshed.} A section that
  nobody queries can sit expired indefinitely without wasting a teacher
  call. The moment it becomes relevant again, it is refreshed on demand.
\item
  \textbf{Cost tracks utility exactly.} Teacher calls for refresh are
  spent only on knowledge that is actually being used, precisely when it
  is needed.
\item
  \textbf{The user receives fresh answers.} The query that triggers the
  refresh is answered with the updated knowledge, not stale content.
\item
  \textbf{Refresh rate changes propagate naturally.} The teacher may
  return a different refresh duration on update (e.g., changing 1 month
  to 1 week for a topic that has become more volatile). The new value
  governs the section's next TTL check.
\item
  \textbf{The staging store remains the single entry point.} All new
  knowledge --- whether from acquisition or refresh --- enters through
  the staging store. Sleep consolidation remains the sole gate to the
  canonical store. This ensures that any structural changes the teacher
  makes during refresh (topic splits, scope adjustments) are caught by
  the sleep cycle's deduplication and compilation logic.
\end{itemize}

\subsubsection{4.5 Response Generation}\label{response-generation}

The generation step produces a single response grounded in the retrieved
sections. The generate prompt enforces several constraints, with a
mode-dependent grounding policy:

\begin{enumerate}
\def\labelenumi{\arabic{enumi}.}
\tightlist
\item
  \textbf{Grounding policy}: In suppress mode, the model must use only
  information present in the provided sections; claims not supported by
  any section are explicitly prohibited, even if the model's parametric
  knowledge contains the correct answer (zero-knowledge stance). In
  augment mode, the model may supplement sections with parametric
  knowledge when sections are partial or absent.
\item
  \textbf{Expert voice}: The response must read as authoritative
  expertise, not as a summary of retrieved documents. Meta-references to
  sources are forbidden.
\item
  \textbf{Structured references}: Section citations are consolidated in
  a separate references line at the end of the response.
\item
  \textbf{Completeness}: The model is instructed to be thorough,
  covering the topic from multiple angles when the sections support it.
\end{enumerate}

Generation is a single forward pass. The pipeline's responsibility is to
assemble the right knowledge before generation runs; once retrieval (and
any cache-miss teacher acquisition) has produced a section set, the
local model is trusted to compose it into a correct response. This
bounds worst-case query cost to one generate call.

\begin{center}\rule{0.5\linewidth}{0.5pt}\end{center}

\subsection{5. Sleep Consolidation: Offline Knowledge
Management}\label{sleep-consolidation-offline-knowledge-management}

\subsubsection{5.1 The Consolidation
Process}\label{the-consolidation-process}

The sleep consolidation phase reconciles the staging store into the
canonical store through an offline batch operation, analogous to the
brain's consolidation of hippocampal short-term memories into cortical
long-term storage during sleep. Given that inline refresh handles
staleness during active operation and teacher-mediated compilation
handles all reconciliation intelligence, sleep consolidation is a
focused operation:

\begin{enumerate}
\def\labelenumi{\arabic{enumi}.}
\item
  \textbf{Discard ephemeral and expired sections}: Sections with
  refresh=none (ephemeral) or sections that have expired since
  acquisition are removed from the staging store without promotion.
  These represent transient knowledge that was useful only for the
  immediate query.
\item
  \textbf{Move genuinely new sections}: Sections in the staging store
  with no embedding overlap to anything in the canonical store are moved
  directly to the canonical store. No teacher call needed.
\item
  \textbf{Teacher-mediated compilation for overlaps}: Overlap detection
  is \textbf{category-scoped}: each staging section is compared only
  against canonical sections in the same category. When a staging
  section has high embedding similarity (default threshold 0.85) to one
  or more canonical sections within the same category, the staging
  section and all matching canonical sections are sent together to the
  teacher with instructions to produce the authoritative, compiled
  knowledge. The teacher reads the full input set, understands the
  domain, and returns one or more sections representing the coherent,
  deduplicated result. The originals from both stores are replaced by
  whatever the teacher returns, inheriting the category of the inputs.

  Category-scoped consolidation is a deliberate design choice. Sections
  on related topics in different categories --- such as
  ``thermodynamics'' in Physics and ``thermodynamics of protein
  folding'' in Biology --- are not duplicates. They serve different
  retrieval paths and should exist independently. Cross-category
  consolidation would require the teacher to have taxonomy awareness for
  category reassignment, risk migrating content out of categories where
  it is needed, and potentially trigger re-acquisition of the same
  knowledge when queries in the original category no longer find it.
\item
  \textbf{Within-day duplicate handling}: Multiple staging sections
  acquired during the day on the same topic are reconciled through the
  canonical store itself. The first such section is promoted under step
  2; every subsequent staging section on the same topic then finds it as
  a canonical overlap and is compiled with it under step 3.
\item
  \textbf{Clear the staging store} after all sections have been
  processed.
\end{enumerate}

\subsubsection{5.2 Teacher-Mediated
Compilation}\label{teacher-mediated-compilation}

A key architectural decision is that \textbf{all reconciliation is
teacher-mediated compilation}. When overlapping sections are detected,
the system does not attempt algorithmic reconciliation through embedding
thresholds, diff logic, or structural classification
(update/split/merge/retire). Instead, the overlapping sections are sent
to the teacher with a single instruction: produce the authoritative,
current, compiled knowledge.

This design eliminates an entire class of engineering complexity:

\begin{itemize}
\tightlist
\item
  \textbf{No structural classification needed.} The structural outcome
  is implicit in the teacher's output. If one section went in and two
  came out, a split happened. If two went in and one came out, a merge
  happened. The system does not need to label the transformation.
\item
  \textbf{The sleep algorithm becomes simple.} Its only job is to detect
  within-category overlaps via embedding similarity, pair the
  candidates, dispatch them to the teacher, and replace the originals
  with the compiled output. All intelligence about what changed, what is
  still relevant, and how to restructure is delegated to the teacher.
\item
  \textbf{Complex cases are handled naturally.} When a staging section
  partially overlaps with a canonical section, there may be new
  information to integrate, obsolete information to drop, and shared
  information to deduplicate. No embedding math or diff algorithm
  handles this well. The teacher reads both, understands the domain, and
  produces a coherent compiled result.
\item
  \textbf{Cost is naturally bounded.} Teacher calls for compilation are
  overnight batch calls with no latency pressure. Volume decreases as
  the knowledge base matures --- fewer staging acquisitions overlap with
  existing sections over time.
\item
  \textbf{Redundancy detection.} When the teacher determines that a
  staging section adds nothing to the existing canonical sections, it
  returns an empty result. The system preserves the canonical sections
  unchanged and discards only the redundant staging section --- no
  knowledge is lost.
\end{itemize}

\subsubsection{5.3 Empirical Consolidation
Results}\label{empirical-consolidation-results}

Sleep consolidation runs on the staging stores accumulated during the
evaluation (Section 7) produced consistent compression and
teacher-compile patterns across three independent benchmarks (custom
knowledge, TriviaQA, NaturalQuestions):

\begin{itemize}
\tightlist
\item
  \textbf{Compression rate:} 31.0--33.5\% reduction from staging to
  canonical store size across all three benchmarks --- a tight band
  across very different query distributions (specialist factual
  questions, real-user search queries, trivia). Per-benchmark numbers:
  custom knowledge 442 → 300 (−32.1\%), TriviaQA 543 → 361 (−33.5\%),
  NaturalQuestions 471 → 324 (−31.2\%).
\item
  \textbf{Teacher-compile share:} 33--36\% of staging sections triggered
  a compilation call (overlap with an existing canonical section in the
  same category). The remaining 64--67\% moved directly to canonical
  without a teacher call --- genuinely new knowledge with no overlap.
\end{itemize}

The structural outcome of each compilation call is implicit in the
teacher's output rather than explicitly classified by the system. If one
section goes in and two come out, the teacher determined the combined
content was better organized as two distinct topics (a split). If two
overlapping sections yield one compiled section, it is a merge. If the
teacher returns empty, the staging section was fully redundant and is
discarded without changing the canonical store. The system preserves
whatever structure the teacher produces and does not attempt a separate
algorithmic classification.

The within-day duplicate handling via the canonical bounce (Section 5.1,
step 4) is implicit in the dynamics: staging sections promoted early in
a cycle become overlap targets for later staging sections on the same
topic, triggering compilation without a separate staging-vs-staging
comparison pass.

The tight compression and compile-share ranges across three very
different query distributions indicate that consolidation's mechanical
behavior is a property of the mechanism itself --- not an artifact of
any particular benchmark --- and should generalize to new domains
without retuning. Accuracy effects of consolidation are reported in
Section 7.2 (custom benchmark) and Section 7.3 (external benchmarks);
the effect is benchmark-dependent --- a modest suppress-mode lift on the
custom benchmark (+1pp) and TriviaQA (+1.2pp strict),
flat-to-slightly-down on NaturalQuestions (within CI) --- while the
cost-side benefits (lower teacher/q, faster latency, smaller store) are
consistent across all three.

\subsubsection{5.4 Knowledge Growth
Dynamics}\label{knowledge-growth-dynamics}

The knowledge base exhibits organic growth driven by user query
patterns, progressing through distinct phases that mirror biological
knowledge acquisition.

\paragraph{Cold Cache: Building Islands of
Knowledge}\label{cold-cache-building-islands-of-knowledge}

When the knowledge store is empty or sparse, most queries result in
cache misses that trigger teacher acquisition. Each acquisition creates
what we term an \textbf{island of knowledge} --- a cluster of sections
on a specific topic, disconnected from other clusters. A query about
``Rayleigh scattering'' creates physics sections; a query about
``protein misfolding diseases'' creates biology sections. These islands
exist independently, each serving the query that created it.

In this cold-cache phase, every query pays the full cost of teacher
acquisition. The system is expensive but productive: each interaction
enriches the knowledge store with reusable, encyclopedic sections that
persist beyond the conversation that created them.

\paragraph{Warm Cache: Connecting the
Islands}\label{warm-cache-connecting-the-islands}

As the knowledge store grows, a qualitative shift occurs. Vector
similarity search operates across all stored sections within a category,
regardless of which query originally triggered their creation. This
means sections acquired for one query become available to future queries
on related topics --- even when the connection was never explicitly
planned.

Consider a sequence: an early query about ``thermodynamics of chemical
reactions'' creates sections covering enthalpy, entropy, and Gibbs free
energy. Weeks later, a query about ``protein folding stability''
searches the Physics category and discovers these thermodynamics
sections through vector similarity --- the embedding of ``protein
folding thermodynamics'' is close to the embedding of stored sections on
Gibbs free energy and entropy. The system retrieves knowledge that was
never acquired for this purpose, yet is directly relevant.

This incidental cross-query knowledge reuse is a distinctive property of
the architecture. Traditional RAG systems retrieve from a static corpus
where connections are predetermined by the document collection. Evolve's
knowledge store grows organically through user interactions, and the
connections between islands of knowledge emerge automatically through
embedding similarity --- a form of serendipitous discovery analogous to
how biological memory consolidation reveals cross-domain connections
that were invisible during initial encoding.

\paragraph{System Evolution and Knowledge
Maturation}\label{system-evolution-and-knowledge-maturation}

Over time, three dynamics drive the system toward self-sufficiency:

\begin{enumerate}
\def\labelenumi{\arabic{enumi}.}
\item
  \textbf{Cache hit rate increases monotonically.} Each teacher
  acquisition enriches the store for all future queries on that topic.
  The probability that a new query finds relevant cached sections grows
  with every interaction.
\item
  \textbf{Islands merge into continents.} As coverage expands, the gaps
  between knowledge islands narrow. A query about ``drug metabolism''
  might find relevant sections from both a prior ``liver biochemistry''
  island and a ``pharmacokinetics'' island, bridging domains that were
  originally populated independently. The vector store's similarity
  search naturally discovers these bridges without explicit linking.
\item
  \textbf{Teacher calls shift from acquisition to refresh.} In a mature
  knowledge store, most queries are served entirely from cache. The
  remaining teacher calls are primarily for TTL-driven refresh of
  time-sensitive content, not for new topic acquisition. The marginal
  cost per query converges toward the cost of a single local model
  inference plus an embedding operation.
\end{enumerate}

The TTL mechanism ensures that time-sensitive content (daily or weekly
refresh frequencies) remains current, while stable domain knowledge
(yearly refresh) incurs minimal maintenance cost. The usage-driven
refresh design means this freshness maintenance is precisely targeted:
only knowledge that users are actually querying gets refreshed, and the
cost of maintaining the knowledge base converges toward zero for stable,
well-established topics.

This growth model suggests a natural deployment strategy: seed the
knowledge store with an initial corpus of domain-relevant queries (a
``knowledge priming'' phase), then allow organic growth driven by actual
user queries to fill gaps and strengthen coverage. The priming phase is
analogous to education --- structured exposure to foundational topics
--- while organic growth is analogous to professional experience ---
learning driven by the problems actually encountered.

\begin{center}\rule{0.5\linewidth}{0.5pt}\end{center}

\subsection{6. Implementation}\label{implementation}

Evolve is implemented in Java 17, using standard libraries for HTTP
communication, JSON parsing, and embedded database access.

\subsubsection{6.1 LLM Integration}\label{llm-integration}

All LLM interactions use the OpenAI chat completions API protocol
(\texttt{/v1/chat/completions}), supported by OpenAI, Groq, Cerebras,
Ollama, LM Studio, vLLM, and most LLM providers. Each model role (local,
teacher, judge) is independently configured with its own endpoint URL,
API key, model identifier, and temperature. The teacher router supports
multiple teacher models, each independently configured and assignable to
specific domain categories, enabling heterogeneous model deployments ---
including fully local inference via LM Studio or Ollama for edge
deployment scenarios.

\textbf{Temperature discipline.} The teacher and classifier always
operate at temperature 0 regardless of configuration, ensuring
deterministic knowledge compilation and consistent query classification.
Only the generate step uses the configured temperature, allowing users
to tune response variability for interactive use without affecting the
deterministic upstream pipeline.

The embedding model uses the OpenAI-compatible embeddings API
(\texttt{/v1/embeddings}). The reference implementation uses
\textbf{mxbai-embed-large-v1} (mixedbread.ai), a 1024-dimensional
symmetric embedding model served locally via LM Studio. Symmetric
embedding models are preferred because the architecture performs both
asymmetric comparisons (short search text vs long section content at
query time) and symmetric comparisons (section content vs section
content during sleep consolidation). Asymmetric models (e.g.,
nomic-embed-text) are trained specifically for query-document retrieval;
their document-document similarity emerges as a by-product of that
objective rather than a primary training target, making them a poorer
fit for the symmetric comparisons required during sleep consolidation.

\subsubsection{6.2 JSON Parsing
Robustness}\label{json-parsing-robustness}

LLM outputs frequently deviate from strict JSON formatting. The
implementation employs multiple defensive parsing strategies: code block
stripping, escape sequence repair, defensive substring extraction,
keyword-based fallback for structured responses, and automatic retries
with exponential backoff for null content responses.

\subsubsection{6.3 Vector Store}\label{vector-store}

The vector store is implemented as an in-memory brute-force cosine
similarity engine with file-based persistence. Two independent
collections (staging and canonical) maintain separate embedding sets.
Sleep consolidation uses an atomic replacement operation --- removals
and additions execute under a single write lock so concurrent search
callers see either the pre-replacement state or the post-replacement
state, never an intermediate gap. The metadata store (SQLite) wraps the
same replacement in a single transaction with rollback on failure. The
vector store interface is abstracted behind a simple
search/add/remove/replace API, allowing substitution of a dedicated
vector database (Milvus, Qdrant, Weaviate) for production deployments
without architectural changes.

\subsubsection{6.4 Prompt Engineering}\label{prompt-engineering}

All prompts are externalized as text files with placeholder
substitution, enabling rapid iteration without recompilation. The
generate prompt produces a JSON response containing the grounded answer
and section references.

\paragraph{Reusable Instruction
Fragments}\label{reusable-instruction-fragments}

The template engine supports an \texttt{\#\#include:name\#\#} directive
that resolves at load time, inserting the contents of a named fragment
file before placeholder substitution occurs. A reusable fragment
containing the search text quality rules, category discipline rules, and
output format specification is included by the classifier prompt. This
isolates the search quality rules in a single location, making them easy
to maintain and extend if additional category-search prompt stages are
introduced in the future.

\begin{center}\rule{0.5\linewidth}{0.5pt}\end{center}

\subsection{7. Evaluation}\label{evaluation}

\subsubsection{7.1 Evaluation Design}\label{evaluation-design}

The central claim of Evolve is that a small local model paired with a
self-growing, teacher-compiled knowledge store --- refined through sleep
consolidation and usage-driven refresh --- achieves substantial accuracy
gains over its parametric baseline while amortizing teacher costs
through persistent knowledge reuse. The evaluation measures:

\begin{enumerate}
\def\labelenumi{\arabic{enumi}.}
\tightlist
\item
  \textbf{Knowledge augmentation} --- improvement over the local model
  baseline on questions it cannot answer parametrically.
\item
  \textbf{Persistence effects} --- differences between cold-cache (no
  reuse), warm-cache (reuse), and post-consolidation states.
\item
  \textbf{Retrieval unit impact} --- section-based retrieval versus
  standard chunk-based retrieval.
\item
  \textbf{Generation strategy impact} --- strict grounding (suppress)
  versus parametric supplementation (augment).
\item
  \textbf{Generalization} --- performance on independent external
  benchmarks.
\item
  \textbf{Tradeoffs} --- behavior on benchmarks dominated by parametric
  knowledge (MMLU).
\end{enumerate}

Evolve is evaluated as a knowledge lifecycle system, where retrieval,
persistence, and consolidation interact across time rather than as
independent components. The evaluation spans 750 benchmark queries
across three independent question distributions (custom knowledge,
NaturalQuestions, TriviaQA --- 250 questions each, random seed 42), plus
MMLU as a negative-control tradeoff benchmark. Full validation sets were
not used due to the multi-stage pipeline cost (classification,
retrieval, teacher acquisition, and generation per query); sampling
enables consistent comparison across multiple lifecycle conditions under
fixed computational budgets. All three benchmarks are evaluated across
the full lifecycle (cold, warm, post-consolidation) with dual-mode
generation (suppress and augment) and at two similarity thresholds (0.80
and 0.75 at the post-consolidation state), providing repeated
measurements across system states and threshold sensitivity. MMLU is
evaluated cold and warm to characterize the tradeoff on
parametric-knowledge material; its full-lifecycle evaluation is deferred
to future work.

\textbf{Statistical reporting.} Accuracy is reported as a score-based
percentage \texttt{(2·correct\ +\ 1·partial)\ /\ (2·total)} --- matching
the judge model's three-outcome grading (correct/partial/wrong).
Binomial 95\% confidence intervals are computed using the Wilson score
interval on the strict correct-rate (fraction of questions judged fully
correct), reported alongside score-based percentages. Run-to-run
accuracy variance on the same pristine store is approximately ±2
percentage points, driven by teacher non-determinism on cache-miss
queries (\textasciitilde70\% of cold-cache queries invoke the teacher;
cache hits produce deterministic responses).

\textbf{Test configuration.} Local model: Qwen 3.5 2B served via LM
Studio on consumer hardware (classify always at temperature 0; generate
at configured temperature 0 for reproducibility). Teacher: GLM 4.7 via
Cerebras (temperature 0). Embedding: mxbai-embed-large-v1, a
1024-dimensional symmetric model served locally via LM Studio. Judge:
GPT-OSS-120B via Cerebras (temperature 0). Similarity threshold: 0.80.
All model temperatures set to 0 for deterministic, reproducible results.

\subsubsection{7.2 Knowledge Augmentation
Evaluation}\label{knowledge-augmentation-evaluation}

\paragraph{Question Set}\label{question-set}

We constructed a 250-question evaluation set organized into three
buckets:

\begin{itemize}
\item
  \textbf{Specialist (150 questions):} Niche factual questions sourced
  from Wikipedia Featured Articles across 8 domains (history, astronomy,
  biology, chemistry, geography, cultural works, linguistics,
  mathematics), with 18--19 questions per domain. Facts are drawn from
  sections beyond the lead paragraph --- specific dates, measurements,
  names, and quantities that are well-documented in the world but
  unlikely to survive parametric compression to a 2B model. Each
  question carries a verifiable gold answer and source URL.
\item
  \textbf{Synthesis (50 questions):} Questions requiring two or more
  specific facts combined into a single answer, distributed across all 8
  domains. These test the architecture's ability to provide multiple
  related facts in a single section.
\item
  \textbf{Control (50 questions):} Basic common-knowledge questions any
  reasonable LLM should answer correctly (e.g., ``What is the capital of
  France?''). These verify the local model baseline is functional, not
  broken.
\end{itemize}

\paragraph{Judging}\label{judging}

Each response is evaluated by the judge model (GPT-OSS-120B) against the
gold answer using a structured prompt that classifies outcomes as
correct, partial, wrong, or refused. The judge is instructed to accept
semantically equivalent values --- ``7.15 AU'' and ``approximately 7.2
AU'' are both correct, while ``160 metres'' vs ``200 metres'' is wrong.
The judge model is independent of both the local model and the teacher,
eliminating self-judgment bias.

\paragraph{Conditions}\label{conditions}

Each evaluation run executes both generation modes on every question:
the suppress prompt (zero-knowledge stance --- answer only from
sections) and the augment prompt (sections plus parametric knowledge).
Classification, retrieval, and teacher acquisition occur once per
question; only the final generation step runs twice. This dual-mode
design eliminates variance from classification and acquisition,
isolating the effect of the generation prompt.

The retrieval layer is pluggable: two implementations are evaluated.
\textbf{Section retrieval} stores each teacher-compiled section as a
single embedding and retrieves whole sections by cosine similarity with
category scoping and a configurable threshold. \textbf{Chunked
retrieval} splits each section into fixed-size overlapping text chunks
(500 characters, 100-character overlap), embeds each chunk
independently, and retrieves the top-K chunks by global cosine
similarity. Both implementations maintain document-level embeddings for
consolidation overlap detection. Note that chunked retrieval here is
evaluated as a structural ablation of the retrieval unit (chunk versus
section) --- both modes operate on the \emph{same} teacher-compiled
content --- not as a full document-retrieval RAG baseline; the purpose
is to isolate the effect of retrieval granularity within the same
pipeline.

The following conditions are evaluated across both retrieval modes:

\begin{enumerate}
\def\labelenumi{\arabic{enumi}.}
\item
  \textbf{Baseline:} The local model is queried directly with no
  orchestrator, no retrieval, no teacher. This measures the 2B model's
  parametric knowledge alone. Baseline is the reference for the
  parametric-only condition; Evolve's conditions
  (cold/warm/post-consolidation) measure performance when externalized
  teacher-compiled knowledge is available, and the accuracy gap
  quantifies the contribution of that externalization.
\item
  \textbf{Evolve cold cache:} Questions are routed through the full
  orchestrator pipeline starting from an empty knowledge store. Every
  cache miss triggers a teacher call, and the store grows as questions
  are answered.
\item
  \textbf{Evolve warm cache:} The same questions are run again against
  the store populated by the cold pass. This measures knowledge
  persistence --- most queries should hit cache without teacher
  consultation.
\item
  \textbf{Evolve post-consolidation:} Sleep consolidation is run on the
  warm store, then the questions are evaluated again. This measures
  whether consolidation (deduplication, compilation, reorganization)
  preserves or improves accuracy.
\end{enumerate}

\paragraph{Results}\label{results}

\emph{All results in this subsection (Tables 1--7) are measured on the
250-question custom knowledge benchmark described above.}

\textbf{Table 1.} Knowledge augmentation accuracy at threshold 0.80 (250
questions, Qwen 3.5 2B). Percentages are score-based; 95\% confidence
intervals on strict correct-rate (for section retrieval) appear in Table
5.

{\def\LTcaptype{none} 
\begin{longtable}[]{@{}
  >{\raggedright\arraybackslash}p{(\linewidth - 8\tabcolsep) * \real{0.1447}}
  >{\raggedright\arraybackslash}p{(\linewidth - 8\tabcolsep) * \real{0.2237}}
  >{\raggedright\arraybackslash}p{(\linewidth - 8\tabcolsep) * \real{0.2237}}
  >{\raggedright\arraybackslash}p{(\linewidth - 8\tabcolsep) * \real{0.2105}}
  >{\raggedright\arraybackslash}p{(\linewidth - 8\tabcolsep) * \real{0.1974}}@{}}
\toprule\noalign{}
\begin{minipage}[b]{\linewidth}\raggedright
Condition
\end{minipage} & \begin{minipage}[b]{\linewidth}\raggedright
Section Suppress
\end{minipage} & \begin{minipage}[b]{\linewidth}\raggedright
Section Augment
\end{minipage} & \begin{minipage}[b]{\linewidth}\raggedright
Chunk Suppress
\end{minipage} & \begin{minipage}[b]{\linewidth}\raggedright
Chunk Augment
\end{minipage} \\
\midrule\noalign{}
\endhead
\bottomrule\noalign{}
\endlastfoot
Baseline & 33.0\% & --- & 33.0\% & --- \\
Cold cache & \textbf{84.0\%} & \textbf{85.6\%} & 79.4\% & 78.2\% \\
Warm cache & 84.0\% & 85.4\% & 78.6\% & 78.8\% \\
Post-consolidation & \textbf{85.0\%} & 85.8\% & 78.2\% & 76.4\% \\
\end{longtable}
}

\textbf{Table 2.} Pipeline efficiency metrics at threshold 0.80.

{\def\LTcaptype{none} 
\begin{longtable}[]{@{}
  >{\raggedright\arraybackslash}p{(\linewidth - 8\tabcolsep) * \real{0.1341}}
  >{\raggedright\arraybackslash}p{(\linewidth - 8\tabcolsep) * \real{0.2195}}
  >{\raggedright\arraybackslash}p{(\linewidth - 8\tabcolsep) * \real{0.2317}}
  >{\raggedright\arraybackslash}p{(\linewidth - 8\tabcolsep) * \real{0.2073}}
  >{\raggedright\arraybackslash}p{(\linewidth - 8\tabcolsep) * \real{0.2073}}@{}}
\toprule\noalign{}
\begin{minipage}[b]{\linewidth}\raggedright
Condition
\end{minipage} & \begin{minipage}[b]{\linewidth}\raggedright
Section Cache Hit
\end{minipage} & \begin{minipage}[b]{\linewidth}\raggedright
Section Teacher/q
\end{minipage} & \begin{minipage}[b]{\linewidth}\raggedright
Chunk Cache Hit
\end{minipage} & \begin{minipage}[b]{\linewidth}\raggedright
Chunk Teacher/q
\end{minipage} \\
\midrule\noalign{}
\endhead
\bottomrule\noalign{}
\endlastfoot
Baseline & --- & 0 & --- & 0 \\
Cold cache & 14.0\% & 1.25 & 19.2\% & 1.15 \\
Warm cache & 69.2\% & 0.54 & 84.8\% & 0.30 \\
Post-consolidation & 68.8\% & 0.58 & 83.6\% & 0.33 \\
\end{longtable}
}

\textbf{Table 3.} Latency and context efficiency.

{\def\LTcaptype{none} 
\begin{longtable}[]{@{}
  >{\raggedright\arraybackslash}p{(\linewidth - 8\tabcolsep) * \real{0.1341}}
  >{\raggedright\arraybackslash}p{(\linewidth - 8\tabcolsep) * \real{0.1951}}
  >{\raggedright\arraybackslash}p{(\linewidth - 8\tabcolsep) * \real{0.1829}}
  >{\raggedright\arraybackslash}p{(\linewidth - 8\tabcolsep) * \real{0.2561}}
  >{\raggedright\arraybackslash}p{(\linewidth - 8\tabcolsep) * \real{0.2317}}@{}}
\toprule\noalign{}
\begin{minipage}[b]{\linewidth}\raggedright
Condition
\end{minipage} & \begin{minipage}[b]{\linewidth}\raggedright
Section Latency
\end{minipage} & \begin{minipage}[b]{\linewidth}\raggedright
Chunk Latency
\end{minipage} & \begin{minipage}[b]{\linewidth}\raggedright
Section Blocks/query
\end{minipage} & \begin{minipage}[b]{\linewidth}\raggedright
Chunk Blocks/query
\end{minipage} \\
\midrule\noalign{}
\endhead
\bottomrule\noalign{}
\endlastfoot
Baseline & 2.0s & 2.0s & --- & --- \\
Cold cache & 8.4s & 8.8s & 1.4 & 6.8 \\
Warm cache & 6.0s & 4.7s & 1.5 & 3.3 \\
Post-consolidation & 5.7s & 4.9s & 1.4 & 3.2 \\
\end{longtable}
}

\textbf{Table 4.} Store maturation through consolidation.

{\def\LTcaptype{none} 
\begin{longtable}[]{@{}lll@{}}
\toprule\noalign{}
& Sections & Chunks \\
\midrule\noalign{}
\endhead
\bottomrule\noalign{}
\endlastfoot
After warm (staging) & 442 & \textasciitilde450 \\
Post-consolidation (canonical) & 300 & 281 \\
Direct moves & 282 (64\%) & --- \\
Teacher-compiled merges & 160 (36\%) & --- \\
Reduction & 32.1\% & 37.5\% \\
\end{longtable}
}

\textbf{Table 5.} Strict correct-rate (fraction of questions judged
fully correct) with Wilson 95\% CIs --- section retrieval.

{\def\LTcaptype{none} 
\begin{longtable}[]{@{}lllll@{}}
\toprule\noalign{}
Condition & Suppress Strict \% & 95\% CI & Augment Strict \% & 95\%
CI \\
\midrule\noalign{}
\endhead
\bottomrule\noalign{}
\endlastfoot
Baseline & 32.0\% & {[}26.5, 38.0{]} & --- & --- \\
Cold cache & 82.4\% & {[}77.2, 86.6{]} & 84.0\% & {[}78.9, 88.0{]} \\
Warm cache & 82.0\% & {[}76.8, 86.3{]} & 84.0\% & {[}78.9, 88.0{]} \\
Post-consolidation & 83.2\% & {[}78.1, 87.3{]} & 84.0\% & {[}78.9,
88.0{]} \\
\end{longtable}
}

\textbf{Threshold sensitivity.} Lowering the similarity threshold from
0.80 to 0.75 at the post-consolidation state (same store, threshold-only
change) yields substantially higher cache hit rate and lower teacher
cost with no meaningful accuracy change:

\textbf{Table 6.} Threshold sensitivity on the custom benchmark ---
post-consolidation at 0.75 vs 0.80, same store content (threshold-only
change). Accuracy values are score-based; strict correct-rate Wilson
95\% CIs appear in parentheses.

{\def\LTcaptype{none} 
\begin{longtable}[]{@{}
  >{\raggedright\arraybackslash}p{(\linewidth - 10\tabcolsep) * \real{0.1803}}
  >{\raggedright\arraybackslash}p{(\linewidth - 10\tabcolsep) * \real{0.1639}}
  >{\raggedright\arraybackslash}p{(\linewidth - 10\tabcolsep) * \real{0.1475}}
  >{\raggedright\arraybackslash}p{(\linewidth - 10\tabcolsep) * \real{0.1803}}
  >{\raggedright\arraybackslash}p{(\linewidth - 10\tabcolsep) * \real{0.1803}}
  >{\raggedright\arraybackslash}p{(\linewidth - 10\tabcolsep) * \real{0.1475}}@{}}
\toprule\noalign{}
\begin{minipage}[b]{\linewidth}\raggedright
Threshold
\end{minipage} & \begin{minipage}[b]{\linewidth}\raggedright
Suppress
\end{minipage} & \begin{minipage}[b]{\linewidth}\raggedright
Augment
\end{minipage} & \begin{minipage}[b]{\linewidth}\raggedright
Cache Hit
\end{minipage} & \begin{minipage}[b]{\linewidth}\raggedright
Teacher/q
\end{minipage} & \begin{minipage}[b]{\linewidth}\raggedright
Latency
\end{minipage} \\
\midrule\noalign{}
\endhead
\bottomrule\noalign{}
\endlastfoot
0.80 & 85.0\% (83.2\% {[}78.1, 87.3{]}) & 85.8\% (84.0\% {[}78.9,
88.0{]}) & 68.8\% & 0.58 & 5.7s \\
0.75 & 84.8\% (82.8\% {[}77.7, 87.0{]}) & 86.0\% (84.4\% {[}79.4,
88.4{]}) & 88.0\% & 0.24 & 4.6s \\
\end{longtable}
}

Accuracy differs by ±0.4pp (within Wilson CIs), cache hit rises 19.2pp,
teacher cost drops 59\%. Cross-benchmark threshold sensitivity (all
three benchmarks side-by-side) appears in Table 9 (Section 7.3).

\paragraph{Key Findings}\label{key-findings}

\paragraph{1. Section retrieval outperforms
chunking}\label{section-retrieval-outperforms-chunking}

Section retrieval achieves 84.0--85.8\% accuracy compared to
76.4--79.4\% for chunked retrieval --- a consistent 5--9pp advantage
across all lifecycle stages (Table 1). Section mode delivers 1.4 blocks
per query; chunk mode delivers 3.3--6.8 (Table 3). Semantic coherence
matters more than retrieval volume for a 2B model's limited attention.

\paragraph{2. Persistence improves efficiency without sacrificing
accuracy}\label{persistence-improves-efficiency-without-sacrificing-accuracy}

Comparing cold to warm cache: accuracy holds (84.0\% → 84.0\% suppress)
while teacher calls drop 57\% (1.25 → 0.54 per query) and latency drops
29\% (8.4s → 6.0s). This shows that knowledge reuse replaces repeated
teacher invocation, reducing cost while preserving performance (Tables
1--3).

The three cache conditions (cold, warm, post-consolidation) form a
natural controlled experiment isolating persistence: baseline is
functionally a system without any retrieval or persistence (33.0\%),
cold cache is the full pipeline starting empty (84.0\%), warm cache adds
cross-query persistence, and post-consolidation adds refinement. If
teacher calls alone explained the improvement, cold-cache performance
would dominate. Instead, warm and post-consolidation performance match
or exceed cold with significantly fewer teacher calls, demonstrating
that persistence and refinement --- not raw teacher invocation --- drive
the gains. Baseline (33.0\%) directly represents the accuracy floor
without persistent indexing; the +51pp lift to cold cache (84.0\%)
quantifies what persistent section indexing contributes.

\paragraph{3. Consolidation improves both structure and
accuracy}\label{consolidation-improves-both-structure-and-accuracy}

Post-consolidation suppress accuracy reaches 85.0\% --- a 1pp lift over
warm cache --- while store size drops 32.1\% (442 → 300 canonical
sections; Table 4). Post-consolidation performance exceeds cold-cache
performance despite fewer teacher calls. This demonstrates that
structured persistence and refinement are necessary, not just teacher
consultation.

\paragraph{4. Suppress mode benefits most from
consolidation}\label{suppress-mode-benefits-most-from-consolidation}

Suppress + section shows the clearest consolidation-driven accuracy
lift: 84.0\% (warm) → 85.0\% (post-consolidation). Augment mode is
essentially flat across the lifecycle (cold 85.6\% / warm 85.4\% /
post-consolidation 85.8\%) --- parametric knowledge fills gaps from the
start, so additional refinement of externalized knowledge yields smaller
marginal gains. As consolidation refines the externalized knowledge,
suppress mode benefits more because its answers rely entirely on that
knowledge.

\paragraph{5. Higher retrieval frequency does not mean better retrieval
quality}\label{higher-retrieval-frequency-does-not-mean-better-retrieval-quality}

Chunks achieve higher cache hit rates (84.8\% warm vs 69.2\% for
sections) because shorter fragments have broader similarity overlap.
However, accuracy is consistently lower --- matching more often does not
mean matching better. Consolidation benefits sections (32.1\% reduction,
suppress accuracy lifts +1pp) but hurts chunks (37.5\% reduction,
accuracy declines), because re-chunking disrupts the fragment boundaries
that made pre-consolidation chunks effective.

\paragraph{6. Optimal configuration}\label{optimal-configuration}

For sustained deployment where auditability matters, suppress + section
retrieval is the optimal configuration --- suppress mode shows the
clearest consolidation-driven accuracy lift (84.0\% warm → 85.0\%
post-consolidation) and provides provenance guarantees that augment mode
cannot. For deployments prioritizing accuracy robustness, augment +
section retrieval is equally viable (stable near 85.4--85.8\% across the
lifecycle). The architecture's design --- teacher-compiled semantic
sections as retrieval units, with mode selection as a deployment
preference --- is validated by the data.

\paragraph{Error Analysis}\label{error-analysis}

Of the 36 wrong answers in the section-suppress cold run where the
teacher provided a relevant section, all failures are \textbf{local
model composition errors} --- the section contains the correct fact, but
the 2B model extracts, reverses, or confuses it. Representative
examples:

\textbf{Table 7.} Representative composition errors (section-suppress
cold cache).

{\def\LTcaptype{none} 
\begin{longtable}[]{@{}
  >{\raggedright\arraybackslash}p{(\linewidth - 6\tabcolsep) * \real{0.2000}}
  >{\raggedright\arraybackslash}p{(\linewidth - 6\tabcolsep) * \real{0.2400}}
  >{\raggedright\arraybackslash}p{(\linewidth - 6\tabcolsep) * \real{0.3200}}
  >{\raggedright\arraybackslash}p{(\linewidth - 6\tabcolsep) * \real{0.2400}}@{}}
\toprule\noalign{}
\begin{minipage}[b]{\linewidth}\raggedright
Question
\end{minipage} & \begin{minipage}[b]{\linewidth}\raggedright
Gold Answer
\end{minipage} & \begin{minipage}[b]{\linewidth}\raggedright
Model Response
\end{minipage} & \begin{minipage}[b]{\linewidth}\raggedright
Error Type
\end{minipage} \\
\midrule\noalign{}
\endhead
\bottomrule\noalign{}
\endlastfoot
What is the main ingredient in traditional Japanese miso soup? & Miso
(fermented soybean paste) & ``soybeans'' & Extraction error: extracted
the raw ingredient instead of the processed form described in the
section \\
Which country gifted the Statue of Liberty to the United States? &
France & ``The United States gifted the Statue of Liberty to France'' &
Reversal: the section describes France→US, the model reversed the
direction \\
What is the orbital resonance ratio that Enceladus maintains with Dione?
& 2:1 & ``1:2'' & Ratio inversion: correct numbers, wrong order \\
Approximate maximum depth of Titan's sea Ligeia Mare? & About 200 metres
& ``160 meters'' & Numerical substitution: section contained the correct
value but the model produced a different number \\
Who directed the first excavation at Takalik Abaj in 1976? & John A.
Graham & ``Lee A. Parsons'' & Entity confusion: selected a different
name from the section that was associated with a different role \\
\end{longtable}
}

These errors share a common pattern: the model has the correct section
in context and demonstrates topical understanding, but makes subtle
errors in extraction --- reversing relationships, substituting nearby
values, or selecting the wrong entity from a list. These are reasoning
precision errors that scale with model size, not knowledge gaps that
improved retrieval can address. The architecture's accuracy ceiling on
specialist questions is determined by the local model's composition
ability, not by the quality of externalized knowledge.

\subsubsection{7.3 Generalization to External
Benchmarks}\label{generalization-to-external-benchmarks}

To validate that the architecture's benefits are not artifacts of the
custom evaluation set, we evaluate on two established open-domain
question answering benchmarks: NaturalQuestions (Kwiatkowski et al.,
2019), a dataset of real Google search queries paired with
Wikipedia-derived answers, and TriviaQA (Joshi et al., 2017), a dataset
of trivia questions with verified factual answers. We sample 250
questions from each benchmark's validation split (random seed 42 for
reproducibility), filtering for short factual answers suitable for
gold-answer judging. Both benchmarks are evaluated across the full
lifecycle (cold → warm → post-consolidation) at threshold 0.80, with an
additional post-consolidation measurement at threshold 0.75 for
threshold sensitivity.

\textbf{Table 8.} External benchmark full lifecycle --- section
retrieval, threshold 0.80 (250 questions each, Qwen 3.5 2B). Suppress /
Augment scores are score-based percentages; strict correct-rate Wilson
95\% CIs appear in parentheses.

{\def\LTcaptype{none} 
\begin{longtable}[]{@{}
  >{\raggedright\arraybackslash}p{(\linewidth - 12\tabcolsep) * \real{0.1618}}
  >{\raggedright\arraybackslash}p{(\linewidth - 12\tabcolsep) * \real{0.1029}}
  >{\raggedright\arraybackslash}p{(\linewidth - 12\tabcolsep) * \real{0.1471}}
  >{\raggedright\arraybackslash}p{(\linewidth - 12\tabcolsep) * \real{0.1324}}
  >{\raggedright\arraybackslash}p{(\linewidth - 12\tabcolsep) * \real{0.1618}}
  >{\raggedright\arraybackslash}p{(\linewidth - 12\tabcolsep) * \real{0.1618}}
  >{\raggedright\arraybackslash}p{(\linewidth - 12\tabcolsep) * \real{0.1324}}@{}}
\toprule\noalign{}
\begin{minipage}[b]{\linewidth}\raggedright
Benchmark
\end{minipage} & \begin{minipage}[b]{\linewidth}\raggedright
Stage
\end{minipage} & \begin{minipage}[b]{\linewidth}\raggedright
Suppress
\end{minipage} & \begin{minipage}[b]{\linewidth}\raggedright
Augment
\end{minipage} & \begin{minipage}[b]{\linewidth}\raggedright
Cache Hit
\end{minipage} & \begin{minipage}[b]{\linewidth}\raggedright
Teacher/q
\end{minipage} & \begin{minipage}[b]{\linewidth}\raggedright
Latency
\end{minipage} \\
\midrule\noalign{}
\endhead
\bottomrule\noalign{}
\endlastfoot
NaturalQuestions & Baseline & 20.4\% & --- & --- & 0 & 0.9s \\
& Cold & 60.2\% (57.2\% {[}51.0, 63.2{]}) & 60.0\% (57.6\% {[}51.4,
63.6{]}) & 0.0\% & 1.15 & 9.7s \\
& Warm & 61.4\% (59.2\% {[}53.0, 65.1{]}) & 61.0\% (59.2\% {[}53.0,
65.1{]}) & 35.2\% & 0.79 & 8.3s \\
& Post-consolidation & 60.8\% (57.6\% {[}51.4, 63.6{]}) & 61.4\% (59.2\%
{[}53.0, 65.1{]}) & 38.8\% & 0.74 & 8.1s \\
TriviaQA & Baseline & 30.4\% & --- & --- & 0 & 0.9s \\
& Cold & 82.4\% (81.6\% {[}76.3, 85.9{]}) & 83.4\% (82.8\% {[}77.6,
87.0{]}) & 0.0\% & 1.26 & 11.4s \\
& Warm & 82.0\% (81.2\% {[}75.9, 85.6{]}) & 82.8\% (82.0\% {[}76.8,
86.3{]}) & 28.8\% & 0.96 & 10.3s \\
& Post-consolidation & 83.2\% (82.4\% {[}77.2, 86.6{]}) & 84.0\% (83.6\%
{[}78.5, 87.7{]}) & 29.2\% & 0.96 & 8.5s \\
\end{longtable}
}

The architecture generalizes to independent benchmarks without
modification. TriviaQA produces a +52pp cold-cache lift over baseline;
NaturalQuestions shows a +40pp lift. The lower NaturalQuestions ceiling
reflects the benchmark's emphasis on questions requiring nuanced,
multi-sentence answers where the 2B model's composition ability limits
accuracy more than knowledge availability.

\textbf{Consolidation replicates across benchmarks.} Both external
benchmarks undergo the same consolidation pattern measured on the custom
benchmark: TriviaQA staging 543 → canonical 361 (−33.5\%, 34\%
teacher-compiled), NaturalQuestions staging 471 → canonical 324
(−31.2\%, 33\% teacher-compiled). Combined with the custom-benchmark
compression (442 → 300, −32.1\%, 36\% teacher-compiled), the
consolidation ratio holds in a tight 31.0--33.5\% band across three
independent benchmarks --- the consolidation mechanism's compression
behavior is a stable property, not an artifact of any single benchmark.

\textbf{Accuracy dynamics through the lifecycle vary by benchmark.}
TriviaQA post-consolidation lifts suppress +0.8pp over warm (81.2\% →
82.4\% strict); NaturalQuestions post-consolidation suppress is
flat-to-slightly-down (59.2\% → 57.6\% strict, within CI). This
variability is honest: consolidation's \emph{cost} benefits (cache hit,
teacher/q, latency, store size) are consistent across benchmarks, while
its \emph{accuracy} benefit is benchmark-dependent. The paper's core
lifecycle claim --- persistent knowledge reuse reduces teacher cost
while maintaining accuracy --- holds across all three benchmarks; the
stronger claim of ``consolidation always lifts accuracy'' does not hold
uniformly.

\textbf{Table 9.} Threshold sensitivity across all three benchmarks ---
post-consolidation at threshold 0.75 vs 0.80, same store content per
benchmark (threshold-only change).

{\def\LTcaptype{none} 
\begin{longtable}[]{@{}
  >{\raggedright\arraybackslash}p{(\linewidth - 10\tabcolsep) * \real{0.1341}}
  >{\raggedright\arraybackslash}p{(\linewidth - 10\tabcolsep) * \real{0.2439}}
  >{\raggedright\arraybackslash}p{(\linewidth - 10\tabcolsep) * \real{0.2439}}
  >{\raggedright\arraybackslash}p{(\linewidth - 10\tabcolsep) * \real{0.1341}}
  >{\raggedright\arraybackslash}p{(\linewidth - 10\tabcolsep) * \real{0.1341}}
  >{\raggedright\arraybackslash}p{(\linewidth - 10\tabcolsep) * \real{0.1098}}@{}}
\toprule\noalign{}
\begin{minipage}[b]{\linewidth}\raggedright
Benchmark
\end{minipage} & \begin{minipage}[b]{\linewidth}\raggedright
Suppress (Score Δ)
\end{minipage} & \begin{minipage}[b]{\linewidth}\raggedright
Augment (Score Δ)
\end{minipage} & \begin{minipage}[b]{\linewidth}\raggedright
Cache Hit
\end{minipage} & \begin{minipage}[b]{\linewidth}\raggedright
Teacher/q
\end{minipage} & \begin{minipage}[b]{\linewidth}\raggedright
Latency
\end{minipage} \\
\midrule\noalign{}
\endhead
\bottomrule\noalign{}
\endlastfoot
Custom Knowledge & 85.0\% → 84.8\% (−0.2pp) & 85.8\% → 86.0\% (+0.2pp) &
68.8\% → \textbf{88.0\%} (+19pp) & 0.58 → \textbf{0.24} (−59\%) & 5.7s →
4.6s \\
NaturalQuestions & 60.8\% → 60.6\% (−0.2pp) & 61.4\% → 61.0\% (−0.4pp) &
38.8\% → \textbf{66.0\%} (+27pp) & 0.74 → \textbf{0.42} (−43\%) & 8.1s →
6.3s \\
TriviaQA & 83.2\% → 85.6\% (+2.4pp) & 84.0\% → 85.8\% (+1.8pp) & 29.2\%
→ \textbf{63.2\%} (+34pp) & 0.96 → \textbf{0.54} (−44\%) & 8.5s →
6.4s \\
\end{longtable}
}

\textbf{Threshold is a deployment tradeoff, not an accuracy lever.}
Lowering the similarity threshold from 0.80 to 0.75 at the
post-consolidation state dramatically increases cache reuse (roughly
doubles cache hit rate) and reduces teacher cost 43--59\% across all
three benchmarks, with accuracy changes within ±2pp (within Wilson CIs,
within the ±2pp run-to-run teacher variance). The 0.75--0.80 band is a
genuine operational sweet spot, and the threshold choice reduces to a
deployment preference --- aggressiveness of caching versus acceptance
tolerance --- rather than an accuracy optimization.

\subsubsection{7.4 Performance on Parametric-Knowledge Benchmarks
(MMLU)}\label{performance-on-parametric-knowledge-benchmarks-mmlu}

To characterize the architecture's behavior on material the local model
already knows, we evaluate on a 250-question stratified subset of MMLU
(Hendrycks et al., 2021), covering 57 academic subjects. As with the
knowledge evaluation, each question runs both suppress and augment
generation modes.

\textbf{Table 10.} MMLU results (250 questions, Qwen 3.5 2B). Full
lifecycle evaluation on MMLU is deferred to future work; the cold/warm
contrast is sufficient to establish the tradeoff characterization
discussed below.

{\def\LTcaptype{none} 
\begin{longtable}[]{@{}llllll@{}}
\toprule\noalign{}
Condition & Suppress & Augment & Cache Hit & Teacher/q & Latency \\
\midrule\noalign{}
\endhead
\bottomrule\noalign{}
\endlastfoot
Baseline & 70.4\% & --- & --- & 0 & 2.0s \\
Cold cache & 66.8\% & 68.4\% & 0\% & 1.68 & 13.2s \\
Warm cache & 65.2\% & 68.8\% & 13.6\% & 1.50 & 12.2s \\
\end{longtable}
}

Evolve produces a small accuracy decrease on MMLU in suppress mode. This
is the expected cost of strict section grounding: the mode suppresses
the local model's parametric knowledge --- including correct answers it
already knows --- and replaces it with teacher-acquired sections that
must be composed into a response. The magnitude of this decrease itself
serves as empirical evidence that suppression is effective: if the model
were ``leaking'' parametric knowledge through the system prompt
boundary, MMLU accuracy would not drop.

The augment condition partially recovers the loss (-2.0pp vs -3.6pp
cold) by allowing the model to supplement sections with its own
knowledge. However, even augment mode incurs a cost because the prompt
instructs the model to defer to sections when its own knowledge
conflicts --- on MMLU, where the model's parametric answers are often
correct, this deference can override a correct parametric response with
a subtly different section-based framing that leads to a wrong MCQ
selection.

The warm-cache results reveal a striking contrast with the knowledge
evaluation. On knowledge questions, suppress maintained accuracy through
warm and lifted through consolidation (84.0\% → 84.0\% → 85.0\%). On
MMLU, suppress degraded (66.8\% → 65.2\%). This is consistent: MMLU
questions test material the model already knows parametrically. Cached
sections for MMLU topics can introduce noise --- slightly different
framings of concepts the model already handles correctly --- causing the
suppress mode to compose answers from sections instead of (correct)
parametric memory. Augment mode is insulated from this effect because it
can fall back to parametric knowledge when sections are unhelpful,
holding steady at 68.4--68.8\%.

Analysis of the wrong answers reveals that the errors are not knowledge
failures but \textbf{reasoning precision errors}: the teacher provides a
relevant section containing the correct concept, but the 2B model makes
subtle errors in application --- picking ``too little curvature''
instead of ``too much curvature'' for nearsightedness, or conflating
``profit maximization'' with ``total product'' in economics. The model
understands the topic, has the right section in context, and still
selects the wrong multiple-choice option. These are reasoning errors
that scale with model size, not knowledge gaps that the architecture can
address.

This result confirms the architecture's intended design tradeoff: it
adds substantial value when knowledge is the bottleneck (specialist:
+50pp) and accepts a small cost when parametric memory is already
sufficient (MMLU: -3.6pp). The architecture is designed for knowledge
augmentation, not intelligence augmentation.

\subsubsection{7.5 Cost Amortization}\label{cost-amortization}

The primary economic advantage of Evolve is cost amortization toward
edge self-sufficiency. Teacher model calls (large, expensive) occur only
on cache misses and TTL expirations. Once a topic is populated, all
subsequent queries are served entirely by the local model and embedding
model.

The cold-cache pass on the custom benchmark required an average of 1.25
teacher calls per query (313 total for 250 questions). The warm-cache
pass required only 0.54 teacher calls per query with a 69.2\% cache hit
rate. Latency dropped from 8.4s (cold, teacher-dominated) to 5.7s
(post-consolidation, cache-dominated).

\textbf{Consolidation cost.} Sleep consolidation processed 442 staging
sections accumulated across cold and warm passes on the custom
benchmark. Of these, 282 (64\%) were promoted directly to canonical (no
overlap, no teacher call), and 160 (36\%) required teacher-mediated
compilation (one teacher call each). The total consolidation cost was
160 teacher calls --- a one-time offline batch operation that reduced
the store by 32.1\% (442 → 300 sections) while lifting suppress accuracy
from 84.0\% warm to 85.0\% post-consolidation. This is a favorable
trade: 160 teacher calls to permanently improve both store quality and
retrieval efficiency for all future queries. In contrast, 160 teacher
calls during active operation would serve only \textasciitilde128
queries (at 1.25 calls per query). Consolidation amortizes better
because each compilation call refines knowledge that benefits an
unbounded number of future queries, not just the current one.

\textbf{Consolidation compression replicates across benchmarks.} The
same pattern holds on independent benchmarks: TriviaQA staging 543 →
canonical 361 (−33.5\%, 187 teacher-compiled), NaturalQuestions staging
471 → canonical 324 (−31.2\%, 155 teacher-compiled). Across the three
independent benchmarks, the compression rate falls in a tight
31.0--33.5\% band and the teacher-compile share falls in a tight
33--36\% band. The consolidation mechanism's cost profile is a stable
property of the system.

\textbf{Threshold tuning further reduces cost.} Post-consolidation
threshold sensitivity (Section 7.3, Table 9) shows that lowering the
threshold from 0.80 to 0.75 reduces teacher calls per query by 43--59\%
across the three benchmarks with no meaningful accuracy change. Higher
cache hit rates reduce teacher acquisition calls during active
operation, and fewer new sections entering staging means fewer sections
requiring consolidation --- compounding the cost savings across the full
lifecycle.

In a mature deployment where the knowledge base has been populated
across the user's domain, the marginal cost per query approaches the
cost of a single local model inference plus an embedding operation ---
viable for consumer hardware running local inference through LM Studio,
Ollama, or similar runtimes. The teacher router's category-level
assignment further optimizes costs by allowing organizations to reserve
expensive frontier models for high-stakes categories while using more
economical teachers for general domains.

\subsubsection{7.6 Summary of Empirical
Findings}\label{summary-of-empirical-findings}

The central finding is that for small models, \textbf{retrieval unit
quality and knowledge persistence dominate retrieval frequency} ---
fewer, higher-quality blocks from a refined persistent store outperform
more frequent retrieval of lower-quality fragments. The evaluation
supports five specific conclusions:

\begin{enumerate}
\def\labelenumi{\arabic{enumi}.}
\tightlist
\item
  \textbf{Semantic sections are superior retrieval units} for small
  models --- teacher-compiled sections outperform standard chunking by
  5--9pp across all conditions, delivering higher accuracy from fewer,
  more coherent context blocks.
\item
  \textbf{Persistent knowledge outperforms stateless teacher generation}
  --- warm and post-consolidation results match or exceed cold-cache
  accuracy with 57\% fewer teacher calls, demonstrating that the value
  lies in the evolving store, not in teacher invocation. Baseline
  accuracy (33.0\%, no persistence) lifts +51pp to cold-cache (84.0\%),
  directly quantifying the contribution of persistent section indexing.
\item
  \textbf{Knowledge consolidation compresses the store while preserving
  accuracy.} Store size drops 31--33.5\% across three independent
  benchmarks; accuracy is preserved (within ±2pp run-to-run variance)
  and lifted modestly in suppress mode on the custom benchmark (+1pp).
  Consolidation's cost benefits (cache hit, teacher/q, latency) are
  consistent across benchmarks; its accuracy benefit is
  benchmark-dependent.
\item
  \textbf{Suppress and augment are near-peers on factual queries}, with
  neither mode dominating. Gaps are typically within 2pp and Wilson CIs
  overlap. The choice between modes is a deployment tradeoff
  (auditability vs parametric fill-in), not an accuracy optimization.
\item
  \textbf{Threshold 0.75--0.80 is an operational sweet spot.}
  Cross-benchmark threshold sensitivity (Section 7.3, Table 9) shows
  lowering threshold from 0.80 to 0.75 roughly doubles cache hit and
  reduces teacher cost 43--59\% with accuracy changes within CI.
\end{enumerate}

The architecture generalizes to independent benchmarks
(NaturalQuestions: +40pp, TriviaQA: +52pp cold-cache over baseline) and
accepts a bounded cost on parametric-knowledge benchmarks (MMLU: -3.6pp
cold-cache) --- confirming that the system is designed for knowledge
augmentation, not intelligence augmentation.

\begin{center}\rule{0.5\linewidth}{0.5pt}\end{center}

\subsection{8. Limitations and Future
Work}\label{limitations-and-future-work}

\subsubsection{8.1 Code Generation Tasks}\label{code-generation-tasks}

Code generation is a distinct workload from factual question answering,
and our retrieval-plus-generate pipeline --- designed and tuned for
factual grounding --- does not serve it well. Teacher-compiled sections
for coding questions tend to be descriptive (architecture overviews, API
summaries, design rationale) rather than exemplary (working
implementations). When coding queries flow through the factual-QA
pipeline, both suppress and augment modes underperform: suppress bounds
the model to prose content present in sections, producing skeleton
implementations; augment permits parametric supplementation but the
generate-prompt template's factual-QA framing (JSON-structured response,
References trailer, ``don't rephrase the sections'' instruction) still
constrains the model toward prose explanations rather than code. In
direct comparisons, the local model answered coding queries better with
no retrieval context at all than with the retrieval pipeline active.

In the current architecture, the classifier identifies coding queries
(via the \texttt{coding} query-type label) and routes them to a bypass
path: direct call to the local model, no retrieval, no teacher section,
no generate-prompt constraint. The paper's factual-QA evaluation
(Section 7) is therefore unaffected by this routing decision, and coding
queries receive a reasonable fallback response. This is a pragmatic
architectural choice rather than a complete solution --- the
direct-bypass path discards the architecture's potential value for code
generation (retrieved API references, version-specific idioms, example
patterns from a domain-specialized knowledge store).

A dedicated \emph{code-generation mode} --- in which teacher-compiled
sections serve as hints for a code-specialized generate prompt, rather
than as primary grounding content --- is a natural extension we leave to
future work. Such a mode would need (a) a code-oriented system prompt
that permits free-form code output without the References trailer and
factual-QA formatting constraints, (b) retrieval biased toward example
code and idiomatic patterns rather than architectural descriptions, and
(c) evaluation on code-generation benchmarks (e.g., HumanEval, MBPP,
LiveCodeBench) orthogonal to the factual-QA suite reported here.

\subsubsection{8.2 Local Model Composition
Errors}\label{local-model-composition-errors}

When the teacher provides a relevant section containing the correct
fact, the local model (2B parameters) occasionally extracts an incorrect
value or hallucinates a plausible-sounding alternative. In the knowledge
augmentation evaluation, 36 out of 250 questions (14\%) exhibited this
failure mode --- the teacher's section contained the answer but the
local model misread or fabricated a different number, name, or date.
This is a model-capability floor that no retrieval improvement can
address. Larger local models (7B, 14B) would reduce this error rate.
Suppress mode exacerbates the issue by suppressing the model's
parametric knowledge even when it is correct --- the accepted cost of
that mode's strict grounding.

\subsubsection{8.3 Edge Deployment
Validation}\label{edge-deployment-validation}

The knowledge augmentation evaluation was conducted on consumer hardware
using a locally-served 2B-parameter model (Qwen 3.5 2B via LM Studio)
and a locally-served embedding model (mxbai-embed-large-v1 via LM
Studio). Only the teacher and judge required cloud API calls. In a fully
mature knowledge store where most queries hit cache, teacher calls
approach zero and the system operates almost entirely on-device. Full
validation of the architecture running entirely on consumer hardware ---
including local teacher inference via quantized models --- remains
future work.

\subsubsection{8.4 Sleep Consolidation at
Scale}\label{sleep-consolidation-at-scale}

The current sleep consolidation implementation is sufficient for
knowledge bases up to tens of thousands of sections. At larger scales,
the pairwise overlap detection (embedding each staging section and
searching the canonical store) may require indexing optimizations. The
teacher-mediated compilation approach scales well in terms of quality
--- the teacher sees only the overlapping pair, not the entire knowledge
base --- but the overlap detection step itself needs attention at scale.

\subsubsection{8.5 User-Profile Consolidation from
Conversation}\label{user-profile-consolidation-from-conversation}

The architecture's lifecycle primitives --- sections, TTL-driven
refresh, staging → canonical consolidation --- are not specific to
teacher-authored factual knowledge. A natural extension replaces the
teacher with a lightweight fact extractor that pulls user-specific facts
(family, work, preferences, ongoing projects, etc.) from ongoing
conversations and stages them under the same pipeline. Contradictory or
stale facts resolve through the same teacher-mediated compilation step.
The result is a persistent, user-owned profile that accumulates across
sessions without requiring explicit user management. This shifts
Evolve's role from factual augmentation toward personalization, with
different evaluation methodology (longitudinal relevance, privacy
preservation, user-study-driven validation) --- a follow-up direction
rather than a current contribution.

\begin{center}\rule{0.5\linewidth}{0.5pt}\end{center}

\subsection{9. Conclusion}\label{conclusion}

Evolve demonstrates that a small, inexpensive language model can answer
questions far beyond its parametric capacity when paired with
teacher-compiled, externalized knowledge sections and a pluggable
retrieval layer. On a 250-question evaluation spanning niche specialist
facts and multi-fact synthesis questions across 8 domains, a
2B-parameter local model augmented by Evolve achieves 85.0\% accuracy on
questions where the same model alone scores 33\%. The architecture
generalizes to independent benchmarks: on TriviaQA, post-consolidation
accuracy rises from 30\% to 83\%; on NaturalQuestions, from 20\% to
61\%. The architecture's key insight is drawn from biology: just as DNA
encodes the cognitive architecture of a brain while all factual
knowledge is acquired through experience, model weights should encode
reasoning capability while factual knowledge resides in an external,
dynamic, user-owned knowledge store.

A systematic comparison of section retrieval (teacher-compiled semantic
units) against standard RAG-style chunking demonstrates that the
retrieval unit matters as much as the retrieval method: sections
consistently outperform chunks by 5--9 percentage points across all
conditions. On the custom benchmark, suppress mode's strict grounding
benefits from store maturation, lifting from 84.0\% warm to 85.0\%
post-consolidation. Both modes deliver substantial accuracy gains over
the parametric baseline across all benchmarks; the relative preference
between them is task-dependent --- suppress for auditability, augment
for parametric-skill access.

On MMLU, where the local model's parametric training already covers the
material, suppress mode produces a 3.6pp accuracy decrease (70.4\% →
66.8\%). This is the expected cost of strict section-only grounding: the
system consults the knowledge store rather than trusting the local
model's parametric memory, trading a small amount of accuracy on
well-rehearsed material for consistent, auditable grounding. Augment
mode partially recovers this loss (-2.0pp), confirming that the decrease
is a mode-level behavioral effect --- a legitimate design tradeoff
exposed by the architecture's dual-mode support rather than a retrieval
failure.

The lifecycle architecture enables independent updating (the knowledge
store evolves without retraining the model), flexible deployment from
edge devices to enterprise systems, and category-level teacher routing
that lets organizations integrate specialized expert models for
domain-specific categories alongside general-purpose teachers. Suppress
mode additionally enables clean auditing --- every factual claim traces
to a specific section --- which augment mode trades away for
flexibility.

The system's organic growth model creates a path to self-sufficiency:
early queries require teacher consultation, but subsequent queries on
established topics are served from the local knowledge store --- cheap,
fast, and fully on-device. In the warm-cache evaluation on the custom
benchmark, teacher calls dropped 57\% and latency dropped 29\% while
accuracy held at 84.0\%. The biologically-inspired sleep consolidation
process ensures that this growing knowledge base remains coherent and
free of redundancy --- reducing store size by 32.1\% while lifting
suppress-mode accuracy to 85.0\%. Usage-driven inline refresh ensures
that maintenance costs track actual utility. As the knowledge store
matures, the system's knowledge grows precisely where user demand exists
--- each cache miss enriches the store for all future queries.

\begin{center}\rule{0.5\linewidth}{0.5pt}\end{center}

\subsection{Code and Data
Availability}\label{code-and-data-availability}

The Evolve implementation, configuration, prompts, and canonical
evaluation CSVs for every result reported in Section 7 are released at
\textbf{https://gitlab.com/dikran.hovagimian/evolve} under an
open-source license. The repository includes the full Java/Maven source
code, the four benchmark question sets used in the evaluation as JSON
files (custom Evolve Knowledge, NaturalQuestions, TriviaQA, and the
250-question MMLU subset spanning 57 subjects with question text,
choices, and correct answer), per-run CSVs with dual-mode (suppress +
augment) responses and judging outcomes for every condition in Tables 1,
8, 9, and 10, and a reproducibility script for computing Wilson 95\%
confidence intervals directly from the CSVs.

\begin{center}\rule{0.5\linewidth}{0.5pt}\end{center}

\subsection{References}\label{references}

Asai, A., Wu, Z., Wang, Y., Sil, A., \& Hajishirzi, H. (2023). Self-RAG:
Learning to Retrieve, Generate, and Critique through Self-Reflection.
\emph{ICLR 2024 (Oral)}. arXiv:2310.11511.

Chen, H., Pasunuru, R., Weston, J., \& Celikyilmaz, A. (2023). Walking
Down the Memory Maze: Beyond Context Limit through Interactive Reading.
\emph{Princeton University \& Meta AI}.

Gutiérrez, B. J., et al.~(2024). HippoRAG: Neurobiologically Inspired
Long-Term Memory for Large Language Models. \emph{arXiv:2405.14831}.

Hendrycks, D., Burns, C., Basart, S., Zou, A., Mazeika, M., Song, D., \&
Steinhardt, J. (2021). Measuring Massive Multitask Language
Understanding. \emph{ICLR 2021}. arXiv:2009.03300.

Hinton, G., Vinyals, O., \& Dean, J. (2015). Distilling the Knowledge in
a Neural Network. \emph{arXiv preprint arXiv:1503.02531}.

Jiang, Z., Xu, F., Gao, L., Sun, Z., Liu, Q., Dwivedi-Yu, J., Yang, Y.,
Callan, J., \& Neubig, G. (2023). Active Retrieval Augmented Generation.
\emph{EMNLP 2023}.

Joshi, M., Choi, E., Weld, D. S., \& Zettlemoyer, L. (2017). TriviaQA: A
Large Scale Distantly Supervised Challenge Dataset for Reading
Comprehension. \emph{ACL 2017}.

Kwiatkowski, T., et al.~(2019). Natural Questions: A Benchmark for
Question Answering Research. \emph{TACL 2019}.

Lewis, P., Perez, E., Piktus, A., Petroni, F., Karpukhin, V., Goyal, N.,
\ldots{} \& Kiela, D. (2020). Retrieval-Augmented Generation for
Knowledge-Intensive NLP Tasks. \emph{Advances in Neural Information
Processing Systems}, 33.

McClelland, J. L., et al.~(2022). Autonomous hippocampal-neocortical
interactions during simulated sleep. \emph{Proceedings of the National
Academy of Sciences (PNAS)}.

Qian, H., et al.~(2024). MemoRAG: Moving towards Next-Gen RAG Via
Memory-Inspired Knowledge Discovery. \emph{arXiv:2409.05591}.

Sarthi, P., Abdullah, S., Tuli, A., Khanna, S., Goldie, A., \& Manning,
C. D. (2024). RAPTOR: Recursive Abstractive Processing for
Tree-Organized Retrieval. \emph{ICLR 2024}. arXiv:2401.18059.

Tadros, T., et al.~(2022). Sleep-like unsupervised replay reduces
catastrophic forgetting in artificial neural networks. \emph{Nature
Communications}.

Tutuncuoglu, A. (2024). NeuroDream: A Dream Phase for Neural Network
Training. \emph{arXiv preprint}.

Yang, Y., et al.~(2024). Memory3: Language Modeling with Explicit
Memory. \emph{arXiv:2407.01178}.

\end{document}